\documentclass[journal]{IEEEtran}
\usepackage{amsmath,amsfonts,amssymb}
\usepackage{algorithmic}
\usepackage{algorithm}
\usepackage{array}
\usepackage[caption=false,font=normalsize,labelfont=sf,textfont=sf]{subfig}
\usepackage{textcomp}
\usepackage{stfloats}
\usepackage{url}
\usepackage{verbatim}
\usepackage{graphicx}
\usepackage{cite}
\usepackage{xspace}
\usepackage{xcolor}
\usepackage{color}
\usepackage{bbding}
\usepackage{lineno}
\usepackage{booktabs}
\usepackage{multirow}
\usepackage{arydshln}
\usepackage{enumitem}
\usepackage{hyperref}

\newcommand{\First}{\textcolor{red}}
\newcommand{\Second}{\textcolor{blue}}

\newcommand{\ie}{\emph{i.e.}}

\begin{document}

\title{Scale-adaptive UAV Geo-Localization \\ via Height-aware Partition Learning}

        % <-this % stops a space
\author{Quan Chen, Tingyu Wang, Rongfeng Lu, Yu Liu, Bolun Zheng and Zhedong Zheng
% <-this % stops a space
% \thanks{Copyright © 2024 IEEE. Personal use of this material is permitted. However, permission to use this material for any other purposes must be obtained from the IEEE by sending an email to pubs-permissions@ieee.org.}
\thanks{Quan Chen is with the School of Automation, Hangzhou Dianzi University, Hangzhou 310018, China, and also with the Key Laboratory of Micro-nano Sensing and IoT of Wenzhou, Wenzhou Institute of Hangzhou Dianzi University, Wenzhou, 325038, China (e-mail:chenquan@alu.hdu.edu.cn).}
\thanks{Rongfeng Lu and Bolun Zheng are with the School of Automation, Hangzhou Dianzi University, Hangzhou 310018, China (e-mail:rongfeng-lu@hdu.edu.cn; blzheng@hdu.edu.cn;).}
\thanks{Tingyu Wang is with the School of Communication
Engineering, Hangzhou Dianzi University, Hangzhou 310018, China (e-mail:tingyu.wang@hdu.edu.cn).}
\thanks{Yu Liu is with the Department of Electronic Engineering, Tsinghua University, Beijing 100084, China (e-mail:liuyu77360132@126.com).}
\thanks{Zhedong Zheng is with the Faculty of Science and Technology, and Institute of Collaborative Innovation, University of Macau, Macau 999078, China (e-mail:zhedongzheng@um.edu.mo).}
\thanks{Tingyu Wang and Bolun Zheng are the Corresponding Authors.}
}

\markboth{Journal of \LaTeX\ Class Files,~Vol.~14, No.~8, August~2021}%
{Shell \MakeLowercase{\textit{et al.}}: A Sample Article Using IEEEtran.cls for IEEE Journals}


\maketitle

\begin{abstract}
UAV Geo-Localization faces significant challenges due to the drastic appearance discrepancy between drone-captured images and satellite views. Existing methods typically assume a consistent scaling factor across views and rely on predefined partition alignment to extract viewpoint-invariant representations through part-level feature construction. However, this scaling assumption often fails in real-world scenarios, where variations in drone flight states lead to scale mismatches between cross-view images, resulting in severe performance degradation. To address this issue, we propose a scale-adaptive partition learning framework that leverages known drone flight height to predict scale factors and dynamically adjust feature extraction. Our key contribution is a height-aware adjustment strategy, which calculates the relative height ratio between drone and satellite views, dynamically adjusting partition sizes to explicitly align semantic information between partition pairs. This strategy is integrated into a Scale-adaptive Local Partition Network (SaLPN), building upon an existing square partition strategy to extract both fine-grained and global features. Additionally, we propose a saliency-guided refinement strategy to enhance part-level features, further improving retrieval accuracy. Extensive experiments validate that our height-aware, scale-adaptive approach achieves state-of-the-art geo-localization accuracy in various scale-inconsistent scenarios and exhibits strong robustness against scale variations. The code will be made publicly available.
\end{abstract}


\section{Introduction}
\IEEEPARstart{U}{nmanned} Aerial Vehicles~(UAVs), also known as drones, have gained popularity due to their ability to efficiently capture data with few occlusions and rich content.
Differ from ground-views, UAVs can easily acquire multi-scale images by changing the flight altitude and camera parameters.
Benefiting from these advantages, UAVs play an irreplaceable role in various fields, including agricultural operations~\cite{goodrich2023placement}, automatic driving~\cite{khan2017uav} and aerial photography~\cite{zhao2023ms}.
Regardless of the applications, precise positioning and navigation of drones are indispensable, which mainly rely on a global navigation satellite system~(GNSS) and high-quality communication environment~\cite{zimmermann2017precise}.
In scenarios with weak or failed signals, GNSS will incorrectly determine the geographic localization of drones.
Therefore, vision-based UAV-view geo-localization~(UVGL) develops a meaningful hot research, aiming to release the dependence on GNSS.
As the key of this task, cross-view matching algorithms calculates the similarity between drone- and satellite-view images to achieve image matching. 
However, due to the appearance difference between drone-views and satellite-views, traditional image matching algorithms~\cite{lin2013cross,bansal2011geo,castaldo2015semantic} are imprecise.

% %*******************************************************
% %*******************************************************
% padding pattern
\begin{figure}[!t]
    \centering
    \includegraphics[width=1.0\linewidth]{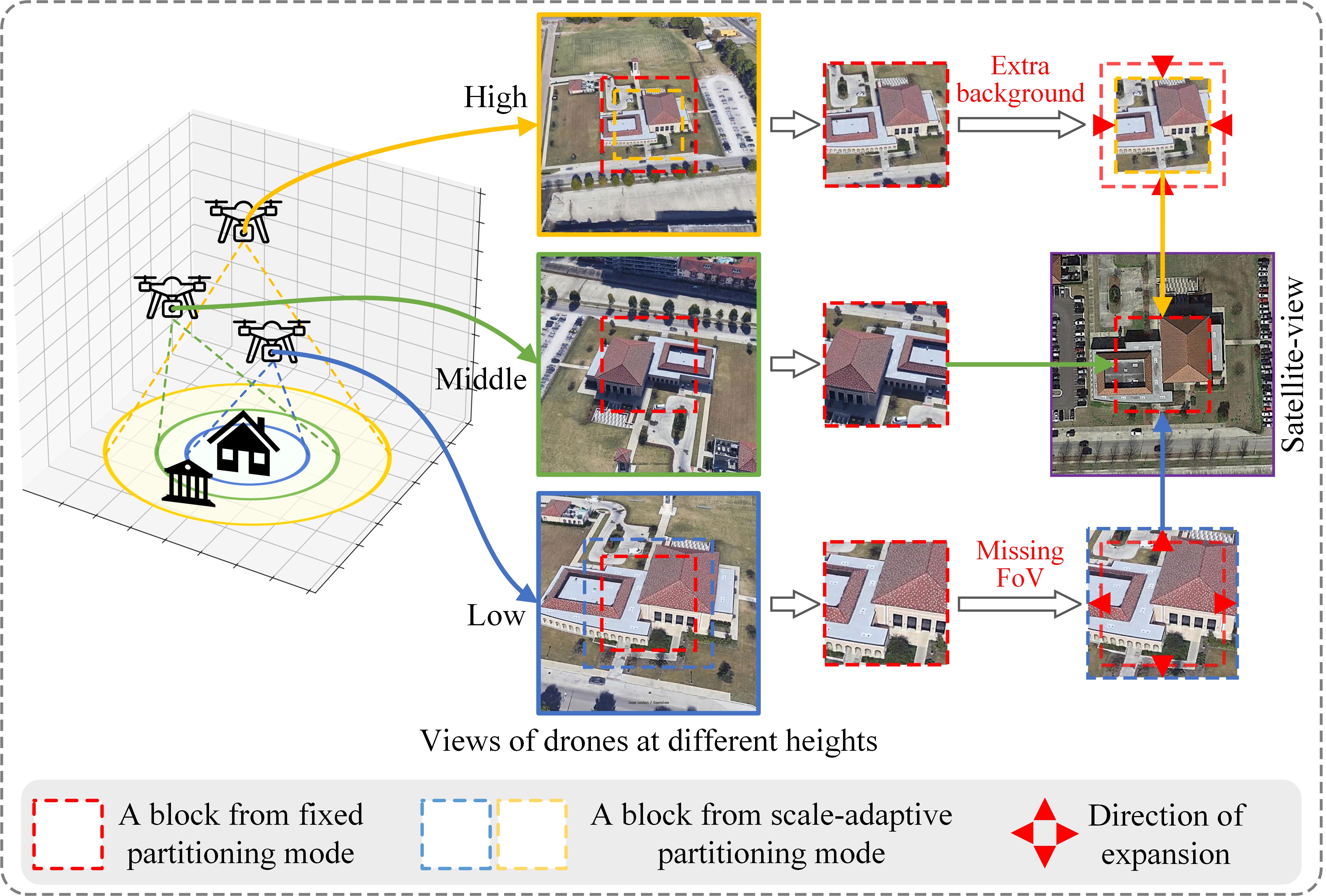}
    \vspace{-.15in}
    \caption{The simplified diagram of our research motivation. Given drone-view images captured from different heights~(high-\textit{yellow box}, middle-\textit{green box}, low-\textit{blue box}) and the corresponding satellite-view image (right), we propose a scale-adaptive partition learning strategy. This strategy dynamically adjusts to different altitudes by eliminating redundant background and expanding the field of view (FoV) to enhance spatial alignment. 
    As shown, the extended/indented partition pairs exhibit more semantically consistent content, facilitating subsequential representation learning.
    }
    \label{motivation}
    \vspace{-.1in}
\end{figure}
% %*******************************************************
% ********************************************************

With the rapid development of deep learning, current methods~\cite{zheng2020university,shen2023mccg,lin2022joint,wang2024learning,wang2024rethinking,zheng2023exif,zheng2024iterated,MCGF,chen2024multi,wang2024multiple,10644040,10636268} utilize pre-trained backbone networks to extract high-level semantic representations and perform image matching based on feature distances, achieving superior accuracy than non-learning-based algorithms.
To obtain robust discriminative representations, plentiful partition learning strategies have been proposed, which construct fine-grained features by segmenting high-level features to fully exploit contextual information.
Depending on the pattern of feature segmentation, partition learning can be divided into two categories: \textbf{soft-partition} learning~\cite{dai2021transformer,zhao2024transfg,li2024geoformer,li2023transformer,zhu2022transgeo} and \textbf{hard-partition} learning~\cite{wang2021each,li2024aerial,ge2024multi,ge2024multibranch,nanhua2024mmhca,lin2024self}.
The former classifies features based on feature values to distinguish the categories of discriminative information, so the shape of partitions is uncontrollable.
FSRA~\cite{dai2021transformer}, as a representative of soft-partition learning, confirms that fine-grained features have a consistent distribution with image content and can boost model performance.
In contrast, inspired by the distribution similarity between drone- and satellite-views, hard-partition strategies set predefined templates with fixed shape to divide high-level features.
As the pioneering hard-partition algorithm, LPN~\cite{wang2021each} designs non-overlapping square-ring templates to build fine-grained features.
Subsequently, several studies~\cite{zhuang2021faster,shao2023style} further optimize the shape and number of templates to facilitate discriminative representation.
Benefiting from the explicit utilization of contextual information that prevents the network from falling into a local optimum, hard-partition strategies reveal superior accuracy and compatibility than soft-partition strategies under same settings.
Nevertheless, hard-partition strategies are more sensitive to spatial changes of input images.
During the data collection process, the variations of drone flight state will lead to the scale mismatch of cross-view images.
The predefined segmentation templates constrain misaligned semantic information to same parts, thus interfering with feature matching.

% %*******************************************************
% %*******************************************************
% padding pattern
\begin{figure}[!t]
    \centering
    \includegraphics[width=1.0\linewidth]{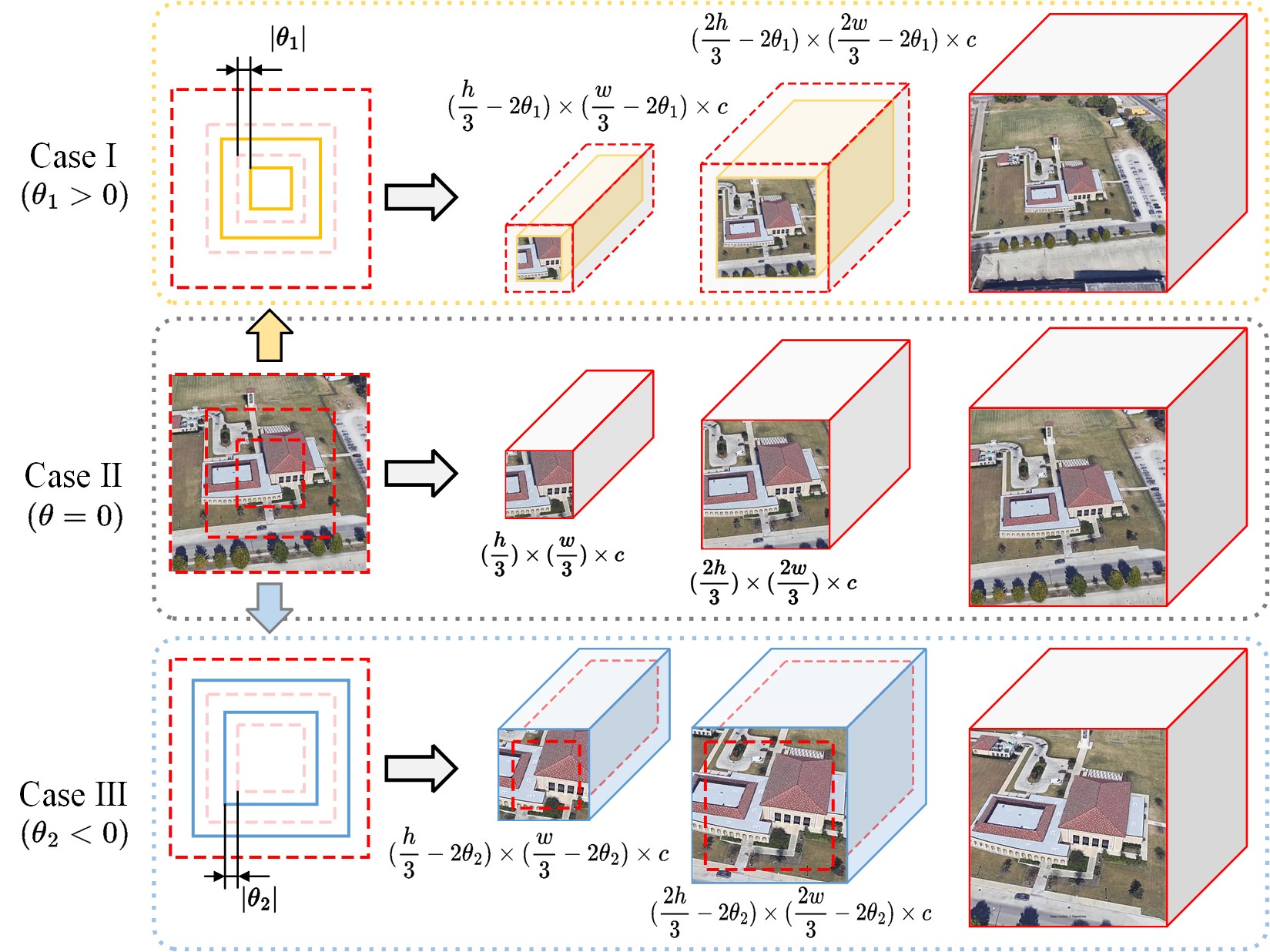}
    \vspace{-.2in}
    \caption{The proposed height-aware adjustment strategy~($N=3$ for illustration).
    Case~(II): Assuming the drone and satellite has similar viewpoint height, we show the typical square partition process, \textit{i.e.}, the uniform partition. 
    However, the assumption is not always hold, and thus, we consider a general form of partition.
    Case~(I): When the drone is with higher flight height with extra backgrounds, the partition areas should be decreased based on the calculated scale factor~$\theta_{1}$.
    Case~(III): When the drone is with lower flight height with limited FoVs, the partition areas should be increased based on the calculated scale factor~$\theta_{2}$.
    }
    \label{dynamic LPN}
    \vspace{-.2in}
\end{figure}
% %*******************************************************
% ********************************************************

% However, there is currently no method to alleviate the dependence of hard-partition learning on cross-view scale consistency.
As shown in Fig.~\ref{motivation}, we depict the limitation of hard-partition strategies in dealing with scale-misaligned image pairs.
% Taking LPN~\cite{wang2021each} as an example, the partition pairs contain similar semantic information in the case of similar view scales, which is conducive to cross-view image matching~(see drone-view with green box).
Taking LPN~\cite{wang2021each} as an example, the partition pairs contain similar semantic information in the case of similar cross-view scales~(see drone-view with green box).
When drone height changes, partition pairs of fixed shape contain misaligned semantic information, defined as imperfect matching~(see drone-view with yellow or blue box).
Following the perspective projection principle~(“closer objects appear larger” in imaging), we argue that adjusting the partition size based on the shooting height of the drone can ensure the content consistency of partition pairs, thus boosting the matching of cross-view partition pairs.
Concretely, for high-height drone-view, indent templates can reduce redundancy in the part-level representations, while expanded templates contributes to the content supplement of partitions for low-height drone-views.
Imitating the process mentioned above, we propose a scale-adaptive partition learning framework named SaLPN, to enhance the content consistency of partition pairs when the shooting height of drone changes.
Specifically, we design a square partition strategy~(SPS) as shown in Case~(II) of Fig.~\ref{dynamic LPN}, which has two typical advantages: 1) multiple partitions constructed by SPS emphasize both fine-grained and global features, thereby exhibiting robustness against scale variations; 2) solid partitions have strong compatibility with subsequent algorithms.
To promote the content consistency of partition pairs, we further propose a height-aware adjustment strategy~(HAAS).
We construct a scale factor $\theta$ based on the relative height ratios between drone and satellite views, and inject it into SPS to dynamically adjust the partition size of drone-view features.
As shown in Case~(I) and Case~(III) of Fig.~\ref{dynamic LPN}, HAAS controls variations of partitions, both in direction and degree.
Furthermore, we design a saliency-guided refinement strategy~(SGRS) to identify targets and environments based on feature activation degrees.
SGRS maintains robustness to scale changes while improving model matching accuracy.
Extensive experiments on University-1652~\cite{zheng2020university} and SUES-200~\cite{zhu2023sues} show that SaLPN achieves state-of-the-art geo-localization accuracy in various scale-inconsistent scenarios and exhibits excellent robustness against scale variations.
For instance, the average R@1 of SaLPN on multiple scale scenarios is 30.81\% and 16.35\% higher than that of Baseline with the instance loss~\cite{zheng2020university} and LPN~\cite{wang2021each}, respectively. Our main contributions are:
\begin{itemize}
    \item We introduce a new scale-adaptive partition learning framework that harness the pre-acquired drone height to  explicitly addresses scale mismatches in cross-view scenarios. In particular, we dynamically adjust partition sizes based on relative height ratios between drone and satellite views, eliminating the conventional assumption of consistent scaling and improving semantic alignment.
    \item We further develop a saliency-guided refinement strategy to enhance the discriminative ability of part-level features by merging the coordinate attention. 
    Specifically, this strategy selectively fuses representations according to the heatmap and extracts the three-level features, \ie, global feature, salient feature and background feature. %addressing a \zznote{xxx} gap in previous approaches. %  thereby boosting the accuracy of geo-localization. 
    \item Extensive experiments show that, in comparison to both hard-partition and soft-partition strategies, our method achieves splendid retrieval performance, especially in scenes with inconsistent cross-view scales, which is common in the real-world drone flight.
\end{itemize}

The rest of the paper is organized as follows:
Section~\ref{Related Works} presents related work, focusing on cross-view geo-localization and part-based representation learning.
In Section~\ref{Methods}, we detail our proposed method, including the method overview and each component.
Section~\ref{Experiments and Results} presents our experimental results, and Section~\ref{Conclusions} provides concluding remarks.

\section{Related Work}\label{Related Works}
We briefly review related works, including cross-view geo-localization and part-based representation learning.

\subsection{Cross-view Geo-localization}
Cross-view Geo-localization aims to recognize the geographic location of an input image based on satellite database, which mainly emphasizes two distinct matching tasks, \textit{i.e.}, the matching of ground and satellite views, as well as the matching of drone and satellite views.
Traditional methods~\cite{lin2013cross,bansal2011geo,castaldo2015semantic} rely on hand-crafted feature, yet their precision is seriously constrained by appearance differences of cross-view images.

Benefiting from the powerful representation ability of neural networks, prevalent methods employ pre-trained backbone networks to extract deep features, thereby enabling cross-view image matching within a high-dimensional space.
Numerous works~\cite{workman2015wide,zhai2017predicting,liu2019lending,deuser2023sample4geo,zhang2023cross,zhu2021vigor,clark2023we} have driven the development of the matching of ground and satellite views in terms of data and algorithms.
For instance, CVUSA~\cite{zhai2017predicting}, CVACT~\cite{liu2019lending} and Vigor~\cite{zhu2021vigor} are extensively utilized as benchmark datasets to evaluate algorithm performance.
Deuser~\textit{et al.}~\cite{deuser2023sample4geo} design GPS-sampling and dynamic similarity sampling to select hard negatives, leading to superior results on several datasets.
To release the cost of collecting precise pairwise data, Li~\textit{et al.}~\cite{li2024unleashing} propose a unsupervised framework to utilize unlabeled data in ground-view geo-localization.
In addition, image synthesis methods~\cite{toker2021coming,shi2022geometry,li2024crossviewdiff} minimizing appearance differences of cross-view images, can improve the matching accuracy.

% %*******************************************************
% %*******************************************************
% padding pattern
\begin{figure*}[!t]
    \centering
    \includegraphics[width=0.95\linewidth]{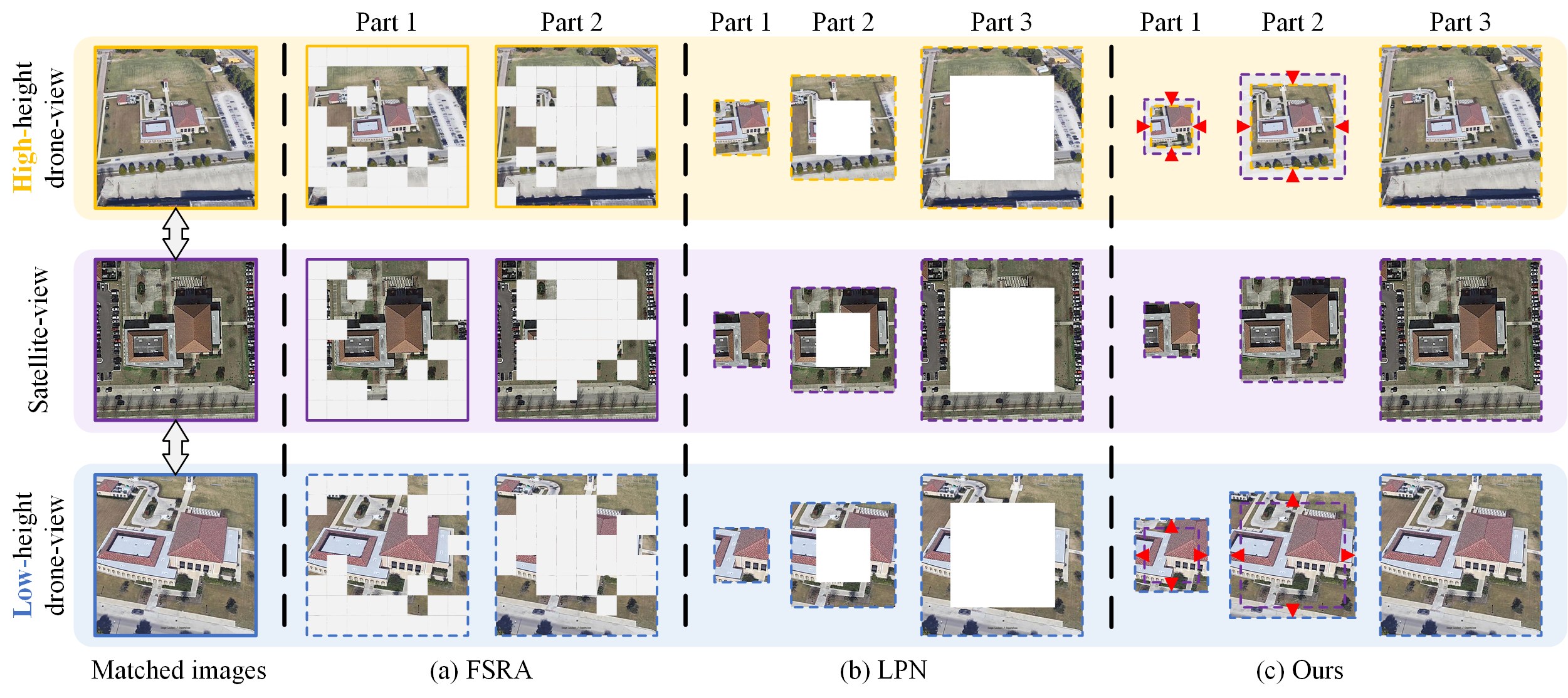}
    % \vspace{-.25in}
    \caption{The comparison of our method with typical partition strategies, including~(a) soft-partition strategy FSRA~\cite{dai2021transformer},~(b) hard-partition strategy LPN~\cite{wang2021each} and (c) our height-aware method.
    The first and third rows are drone-view images with higher and lower heights, while the second row is the matched satellite image with a single scale. For different partition strategies, the corresponding part-level image representations are placed in the same column. We could observe that the proposed method has yielded a more consistent inter-part alignment at the same part level (\ie, every column).}
    \label{different partitions}
    \vspace{-.1in}
\end{figure*}
% %*******************************************************
% ********************************************************

With the growing popularity of UAV devices, several datasets~\cite{zhu2023sues,dai2023vision,ji2024game4loc} introduce drone-view images to support for the UVGL task.
Zheng~\textit{et al.}~\cite{zheng2020university} collect the University-1652 dataset, which contains pairs of ground-views, drone-views and satellite-views. 
Notably, Zheng~\textit{et al.}~\cite{zheng2020university} introduce variations in both the shooting angle and shooting height of drone-views, thus enhancing the diversity of University-1652.
To investigate effects of drone-views at various heights on geo-localization, Zhu~\textit{et al.}~\cite{zhu2023sues} develop SUES-200 comprising drone images acquired at four altitudes.
DenseUAV~\cite{dai2023vision}, as the pioneering low-altitude urban scene dataset, is tailored for the drone self-positioning task.
Focusing on inaccurate alignment between query images and corresponding labels, GTA-UVA~\cite{ji2024game4loc} introduces semi-positive samples to simulate imperfect matching.
Despite viewpoint changes compared to ground-view geo-localization, mainstream solutions~\cite{zheng2020university,shen2023mccg,lin2022joint,wang2024learning,ye2024coarse,li2024learning,he2024contrastive,sun2023f3,wang2024sequence,liu2024segcn} converge upon a similar principle: mapping multi-view images into a shared high-dimensional space, subsequently facilitating image matching through similarity calculation.
Several studies~\cite{dai2021transformer,li2024aerial,10587023} prove that stronger backbone networks can significantly improve the accuracy of UVGL.
Sun~\textit{et al.}~\cite{sun2023f3} design a specific feature extractor named F3-Net to capture both local and global information.
RK-Net~\cite{lin2022joint} inserts unit subtraction attention modules inside the backbone to detect representative keypoints, yielding robust features against viewpoints. 
Considering the color discrepancy between drone-views and satellite-views, recent methods~\cite{tian2021uav,sun2024tirsa} adjust the color distribution of drone-views to reduce domain gaps.
Some efforts~\cite{zhao2024transfg,he2024contrastive} improve the optimization function in the training phase to facilitate domain alignment of cross-view images.
To obtain robust semantic representations, plentiful partition learning strategies are also introduced, which will be detailed in next subsection.
In this paper, we propose a UAV-view geo-localization algorithm applicable to scenes with inconsistent cross-view scales.
To meet the requirements of scale inconsistency and scene diversity, we conduct experiments on University-1652~\cite{zheng2020university}.

\subsection{Part-based Representation Learning}
Fine-grained features guide models to capture more comprehensive information, which has been proven effective in many fields, such as image recognition~\cite{shen2023pedestrian,shen2023triplet,chen2019destruction} and segmentation~\cite{sun2024ultrahighresolutionsegmentationboundaryenhanced,cheng2024sptsequenceprompttransformer,yin2024fine}.
Inspired by classical part-based descriptors~\cite{ojala2002multiresolution,lowe1999object,bouchard2005hierarchical}, numerous works in person re-identification~\cite{xu2018attention,zhong2019invariance,song2019generalizable,sun2018beyond,li2017learning,zheng2022parameter} develop local feature representation by introducing human body structure information.
For instance, OG-Net~\cite{zheng2022parameter} leverages 3D body keypoints to capture consistent body features from 2D images, thus enabling the alignment of macro- and micro-body features across images.
To eliminate the dependence on skeleton points, researchers further propose coarse partition strategies rooted in the vertical distribution of the human body~\cite{sun2018beyond,li2017learning,shen2024stepnet}.

Previous research has laid the foundation for partition learning in UVGL, which can be divided into two categories based on the pattern of feature segmentation: \textbf{soft-partition} learning~\cite{dai2021transformer,zhao2024transfg,li2024geoformer,li2023transformer} and \textbf{hard-partition} learning~\cite{wang2021each,li2024aerial,ge2024multi,ge2024multibranch,nanhua2024mmhca}.
Soft-partition strategies, represented by FSRA~\cite{dai2021transformer}, leverage feature values to distinguish categories of semantic information.
Subsequently, TransFG~\cite{zhao2024transfg} adopts the second-order gradient of features as the partition index, yielding a decent performance boost.
FSRA and TransFG divide features equally based on the setting that each part has equal area, resulting in misalignment of cross-view partition pairs in both large-scale and small-scale scenes.
To address this issue, GeoFormer~\cite{li2024geoformer} utilizes a k-modes clustering algorithm to divide features adaptively.
However, the above soft-partition strategies rely on the transformer framework~\cite{dosovitskiy2020image}.
Drawing upon the similarity between drone- and satellite-views,  some methods employ fixed templates to generate fine-grained features, which are defined as hard-partition strategies.
LPN~\cite{wang2021each} designs non-overlapping square-ring templates to segment features, thus mining contextual information.
Based on LPN, several studies~\cite{zhuang2021faster,shao2023style,nanhua2024mmhca,ge2024multibranch} optimize the shape and number of templates to further facilitate discriminative representation.
Aiming at the problem of position offset, SDPL~\cite{10587023} proposes a shifting-fusion strategy, which adjusts the segmentation center to align partitions with non-centered targets.
The hard partition strategies exhibit commendable accuracy and compatibility, while being sensitive to cross-view scale variation.
Recently, Safe-Net~\cite{lin2024self} proposes a saliency-guided self-adaptive partition strategy, which achieves preferable results than previous methods in scale inconsistency scenarios.
However, Safe-Net~\cite{lin2024self} also produces significant performance degradation in extreme scenarios.
We intend to design an robust algorithm against scale variations for UVGL.

To visually reveal the dependence of partition learning on cross-view scales, we illustrate two representative methods~(FSRA~\cite{dai2021transformer} and LPN~\cite{wang2021each}) for dealing with high-altitude scenes and low-altitude scenes, as shown in Fig.~\ref{different partitions}.
Limited by the setting of equal partition area, FSRA divides features of different categories into the same partition, resulting in feature misalignment.
LPN employs fixed templates to generate fine-grained features.
Notably, in scenarios where cross-view scales are inconsistent, partition pairs contain misaligned semantic content.
In this paper, we study a part-based representation learning for UVGL. 
In comparison with soft-partition strategies, our method achieves superior performance and compatibility, whereas, in contrast to hard-partition strategies, it exhibits enhanced robustness against scale variations.

\section{Methodology}\label{Methods}

\textbf{Task definition.}
The geo-localization dataset contains image pairs captured from different platforms, defined as~$\{x_{D},x_{S}\}$, where subscripts $D$ and $S$ indicate drone and satellite-platform, respectively.
We denote the label as $y\in [1, C]$, where $C$ indicates the number of categories.
For instance, the label of the University-1652~\cite{zheng2020university} dataset is $y\in [1, 701]$.
For UAV-view geo-localization, we intend to learn a mapping function that projects images from various platforms to one shared semantic space.
Image pairs with same labels are close to each other, and vice versa are far away. Therefore, given the drone-view images, user could quickly search the corresponding images from other platform during the inference time. According to the meta-data of other platform, we localize the drone even in the GNSS-denied areas. 

\begin{figure*}[!t]
    \centering
    \includegraphics[width=0.95\linewidth]{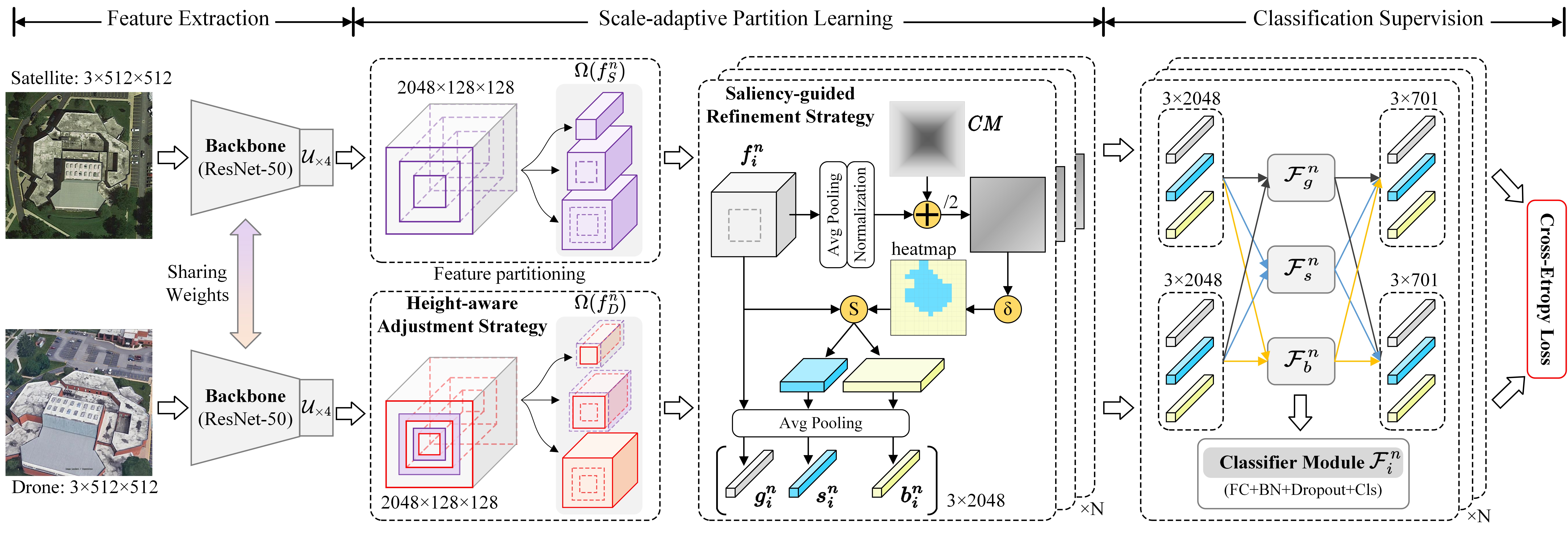}
    \caption{Overview of SaLPN framework, including three phase: feature extraction, scale-adaptive partition learning and classification supervision. 
    In the feature extraction phase, we extract the visual features by the backbones with sharing weights between two platforms. 
    In the scale-adaptive partition learning phase, the visual features from two branch are sliced into part-level features with semantically consistent content. 
    Next, each part-level feature is refined into more fine-grained feature descriptors, including global, salient and background representations, via a saliency-guided refinement strategy.
    In the classification supervision phase, we leverage the classifier module to predict the geo-tag of all feature descriptors. The network is optimized by minimizing the sum of the cross-entropy losses over all parts. 
    In the testing phase, part-level image representation is extracted before classification layer in classifier module, and measures similarity by Euclidean distance.
    \textcircled{\scalebox{0.8}{+}} denotes the element-wise addition;
    \textcircled{\scalebox{0.8}{$\delta$}} denotes threshold-based binarization;
    \textcircled{\scalebox{0.8}{S}} denotes the feature separation.
    }
    \label{framework}
\end{figure*}

\textbf{Overview.} As depicted in Fig.~\ref{framework}, SaLPN consists of three stages: feature extraction, scale-adaptive partition learning and classification supervision.
In the feature extraction stage, a backbone network maps drone- and satellite-view images to high-level feature maps.
Then, we introduce scale-adaptive partition learning to generate fine-grained features with consistent content.
We further propose a saliency-guided refinement strategy to classify each part-level feature, thus obtaining finer discriminative representations.
Finally, we adopt a classifier to predict the category of features obtained from different views, followed by using cross-entropy loss to optimize the model.

% %*******************************************************
% %*******************************************************
% padding pattern
% %*******************************************************
% ********************************************************

\subsection{Feature Extraction}\label{Feature Extraction}
Both drone-view and satellite-view images fall within the category of aerial-view images, exhibiting comparable feature domains. 
Consequently, we employ one feature extractor to process cross-view images.
As mentioned before, SaLPN belongs to hard-partition learning and is compatible with various backbone networks, such as ResNet~\cite{he2016deep} and ViT~\cite{dosovitskiy2020image}.
Following previous works~\cite{wang2021each,10587023}, we adopt ResNet-50 for illustration.
Given an image $x_{i}\in\mathbb{R}^{3\times512\times512}$, after the backbone network and an upsampling layer, we can acquire the corresponding high-level feature map $f_{i}\in\mathbb{R}^{2048\times128\times128}$. 
The process can be formulated as:
\begin{equation}\label{eq1}
\left\{\begin{array}{c}
    f_{D}=\mathcal{U}_{\times4}(\mathcal{F}_{backbone}(x_{D})),\\
    f_{S}=\mathcal{U}_{\times4}(\mathcal{F}_{backbone}(x_{S})),
\end{array}\right.
\end{equation}
where $f_{D}$ and $f_{S}$ are the high-level features of the drone-view and satellite-view, respectively; $\mathcal{F}_{backbone}$ stands for the process of feature extraction; $\mathcal{U}_{\times4}$ stands for a $\times4$ up-sampling layer.
We emphasize that the degree of variation in partition size can be refined by upsampling feature maps.

\subsection{Scale-adaptive Partition Learning}\label{Distance Guided Dynamic Partition Learning}
To capture aligned contextual information even in scale-inconsistent scenes, we propose a scale-adaptive partition learning, consisting of a square partition strategy, a height-aware adjustment strategy, and a saliency-guided refinement strategy.

\textbf{Square partition strategy~(SPS).}
We propose a simple yet effective partition strategy to generate fine-grained features.
As shown in Case(II) of Fig.~\ref{dynamic LPN}, SPS splits the input feature into $N$ parts following square templates.
Concretely, after SPS processing, the high-level feature map $f_{i}\in\mathbb{R}^{2048\times128\times128}$ is segmented into $N$ part-level feature $f_{i}^{n}\in\mathbb{R}^{2048\times \frac{128 n}{N} \times \frac{128 n}{N}}(n\in[1,N])$, where $n$ denotes the $n$-th part. The conventional SPS processing can be formulated as:
\begin{equation}\label{eq2}
    \Omega(f_{i}^{n}) = \{f_{i}^{n}|\mathcal{F_{SPS}}(f_{i};N)\},
\end{equation}
where $\mathcal{F}_{SPS}(\cdot)$ stands for the process of SPS, and $\Omega(f_{i}^{n})$ denotes the set of all part-level features.
Despite the similarities with LPN, comparing Fig.~\ref{different partitions}~(b) and~(c) shows that SPS mines local features while maintaining global information, enhancing robustness against scale changes.

\textbf{Height-aware adjustment strategy~(HAAS).}
Existing hard-partition methods adopt uniform templates for processing multi-view images.
However, in scenarios where drone-view scale undergo significant variations, these methods exhibit notable performance degradation. 
The primary reason is the misalignment of semantic content within partition pairs.
To alleviate this limitation, we propose an innovative height-aware adjustment strategy,~\textit{i.e.}, dynamically adjusting partition size in response to drone-view scale changes, thus boosting content consistency between partition pairs.

First, we assume that the shooting height of drone-views is acquirable, while the height of satellite-view images can be estimated by comparing the area per unit pixel with a fixed height drone-view image.
To quantify the change in partition size, we further construct a scale factor $\theta$ based on the relative height between cross-view images:
\begin{equation}\label{eq3}
    \theta = \text{round}(\frac{H_{D}-H_{S}}{H_{S}} \cdot \alpha),
\end{equation}
where $H_{D}$ and $H_{S}$ denote the height from drone-view and satellite-view to the ground, respectively; $\text{round}(\cdot)$ stands for the rounding operation, and partition adjustment factor $\alpha$ denotes a hyper-parameter utilized to regulate the degree of partition change. 
($\theta$$>$0) indicates that the partition needs to be decreased, and ($\theta$$<$0) vice versa.
% The sign of $\theta$ is determined by $H_{D}$, where~($\theta$$>$0) denotes indenting partitions, whereas~($\theta$$<$0) denotes expanding partitions.
Note that the height of satellite-view $H_{S}$ is constant, which is the regular setting for UVGL.

Then, different from the traditional SPS in Eq.~\ref{eq2}, we further take $\theta$ into consideration, which adjusts partitions of drone-views, which can be formulated as:
\begin{equation}\label{eq4}
    \Omega(f_{D}^{n}) = \{f_{D}^{n}|\mathcal{F_{HAAS}^{D}}(f_{D};N;\theta)\},
\end{equation}
where $\mathcal{F_{HAAS}^{D}}(\cdot)$ denotes the process of HAAS specific for drone-views. 
For ease of understanding, we depict the $\mathcal{F_{HAAS}^{D}}$ process in two cases~(\textit{i.e.}, $\theta$$>$0 and $\theta$$<$0), as shown in Fig.~\ref{dynamic LPN}~(Case~I and~III).
We employ $\theta$ to adjust the size of partitions, and the shape of each partition pair is expressed as follows:
\begin{equation}\label{eq5}
\left\{\begin{array}{l}
    f_{D}^{n}\in \mathbb{R}^{2048\times(\frac{128n}{N}-2\theta)\times(\frac{128n}{N}-2\theta)},\\
    f_{S}^{n}\in \mathbb{R}^{2048\times(\frac{128n}{N})\times(\frac{128n}{N})}.
\end{array}\right.
\end{equation}

% todo
In practice, there are two extreme cases that should be emphasized: 1) when $\theta$$>$0, the size of the smallest partition must be positive; 2) while $\theta$$<$0, the minimum partition block expands into global features, \ie, all partitions are equal to the global feature. 
This means that $\alpha$ in Eq.~\ref{eq3} has a limited range when the height range of the dataset is known. 
Next, we will derive the range of values of $\alpha$.

% \theta>0 
Specifically, when $\theta$$>$0~(\ie, $H_D$$>$$H_S$), we have $1 \leq \frac{128}{2N}-\theta$ to avoid non-positive partitions. Based on Eq.~\ref{eq3}, we can derive the range of the hyperparameter $\alpha$:
\begin{equation}\label{eq6}
    0\leq \alpha \leq \frac{64-N}{N}\cdot \frac{H_S}{H_D^{max}-H_S},
\end{equation}
where $H_D^{max}$ denotes the maximum height of drone-views.

% \theta<0 
Conversely, when $\theta$$<$0~(\ie, $H_D$$<$$H_S$), we have $\frac{128}{2N}-\theta\leq\frac{128}{2}$. Based on Eq.~\ref{eq3}, we also can derive the range of the hyperparameter $\alpha$:
\begin{equation}\label{eq7}
    0\leq \alpha \leq \frac{64(N-1)}{N}\cdot \frac{H_S}{H_S-H_{D}^{min}},
\end{equation}
where $H_D^{min}$ denotes the minimum height of drone-views.

If all same-height image pairs have $H_S = H_D$, we could derive $\theta = 0$, which is independent from the selection of $\alpha$. 
Therefore, for a dataset with a clear height range, the value of $\alpha$ needs to satisfy both Eq.~\ref{eq6} and \ref{eq7}.
% In the ablation studies, we also observe that we can select $\alpha$ from a large range, and the model is not very sensitive to the selection of  $\alpha$.

\textbf{Saliency-guided refinement strategy~(SGRS).}
Each solid partition inherently contains both target and environment information. 
To further refine part-level features, we design a saliency-guided refinement strategy.
The overall process is depicted in Fig.~\ref{framework}, and can be expressed as:
\begin{equation}
   (g_{i}^{n},s_{i}^{n},b_{i}^{n}) = \mathcal{F}_{SGRS}^{n}(f_{i}^{n}),
\end{equation}
where $\mathcal{F}_{SGRS}^{n}(\cdot)$ denotes the process of SGRS, and $g_{i}^{n}$, $s_{i}^{n}$, $b_{i}^{n}$ are denote the output global feature, salient feature and background feature, respectively.

Specifically, assuming the input partition feature is $f_{i}^{n}\in\mathbb{R}^{2048\times h\times w}$, we first generate the heatmap $heat_{i}^{n}\in\mathbb{R}^{1\times h\times w}$ to reflect importance regions of the input.
Considering that targets are usually centered, we introduce an auxiliary coordinate map~($CM$) to direct the heatmap's focus towards the central region.
$CM$ indicates inverse pixel-to-center distance, ranging from 1~(center) to 0~(farthest).
To be consistent with the partition shape, we adopt Chebyshev distance to generate $CM$:
\begin{equation}
CM(u,v)= max(\frac{h}{2},\frac{w}{2}) - max(|u-\frac{h}{2}|,|v-\frac{w}{2}|),
\end{equation}
% \zznote{using u,v. p is probablity.} 
where $(u,v)$ denotes the coordinates of pixels.
Note that other distance metrics, \textit{e.g}, Euclidean distance and Manhattan distance, will be compared in the experiments.
The generation of $heat_{i}^{n}$ can be formulated as:
\begin{equation}
    heat_{i}^{n}(u,v) = (Norm(Avg(f_{i}^{n}(u,v) )) + CM(u,v))/2,
\end{equation}
where $Norm(\cdot)$ denotes the normalization operation, and $Avg(\cdot)$ denotes the average pooling operation.
Next, we set a threshold $\delta$ to split the heatmap, resulting in a binary mask $Mask_{i}^{n}$:
\begin{equation}
    Mask_{i}^{n}(u,v) = \left\{\begin{array}{ll} 1, & heat_{i}^{n}(u,v)\geq \delta, \\
    0, &  heat_{i}^{n}(u,v)<\delta.
    \end{array}\right.
\end{equation}
Finally, in reference to FSRA~\cite{dai2021transformer}, we leverage the value of $Mask_{i}^{n}$ to split the input feature $f_{i}^{n}$, which can be expressed as follows:
\begin{equation}\label{eq13}
\left\{\begin{array}{l}
    s_{i}^{n} = Avg(f_{i}^{n}(u,v) \odot Mask_{i}^{n}(u,v) ),\\
    b_{i}^{n} = Avg(f_{i}^{n}(u,v) \odot (1-Mask_{i}^{n}(u,v))),\\
    g_{i}^{n} = Avg(f_{i}^{n}),
\end{array}\right.
\end{equation}
where $\odot$ denotes the pixel-level multiplication, $s_{i}^{n}\in\mathbb{R}^{2048\times1\times1}$ and $b_{i}^{n}\in\mathbb{R}^{2048\times1\times1}$ denote the salient feature and background feature, respectively.
We also extract $g_{i}^{n}\in\mathbb{R}^{2048\times1\times1}$ to preserve the input information.
It is worth mentioning that SGRS is implemented for each part-level features of both drone-view and satellite-view, resulting in $3N$ feature vectors.

\subsection{Classification Supervision}\label{Classification supervision}
We employ 3N classifier modules to map feature vectors of all sources into three corresponding shared space. 
As shown in Fig.~\ref{framework}, the classifier module includes a fully connected layer~(FC), a batch normalization layer~(BN), a dropout layer~(Dropout), and a classification layer~(Cls). The process can be expressed as:
\begin{equation}
   p_g^{i,n} =\mathcal{F}_{g}^{n}(g_{i}^{n}),\quad
   p_s^{i,n} =\mathcal{F}_{s}^{n}(s_{i}^{n}),\quad
   p_b^{i,n} =\mathcal{F}_{b}^{n}(b_{i}^{n}),
\end{equation}
where $\mathcal{F}_{g}^{n}, \mathcal{F}_{s}^{n}, \mathcal{F}_{b}^{n}$ are three independent classifier modules, respectively.
It is worth noting that the classifiers adopted for different granularity feature vectors are also independent.
Subsequently, we adopt a cross-entropy loss function to optimize our framework, with the objective of minimizing the distance between features of the same geo-tag:
\begin{equation}
    \mathcal{L}_{CE} = \sum_{i,n} -log(p_g^{i,n}(y))  -log(p_s^{i,n}(y)) -log(p_b^{i,n}(y)),
\end{equation}
where $p_g(y)^{i,n}$ denotes the logit score of the ground-truth geo-tag $y$ based on the global feature $g^n_i$. 
Similarly, $p_s^{i,n}(y)$ and $p_b^{i,n}(y)$ are the predicted probability of the ground-truth location based on the salient and background feature, respectively. 
%It is noteworthy that the losses on the image of different features, \ie, global feature, salient feature and background feature, are calculated independently.

%%%%%%%%%%%%%%%%%%%%%%%%%%%%%%%%%%%%%%%%%%
\section{Experiment}\label{Experiments and Results}
To evaluate the effectiveness of SaLPN for UAV-view geo-localization, we conduct comparative experiments on the University-1652 and SUES-200 datasets, and ablation experiments on the University-1652 dataset.  
These experiments assess the performance of our method in terms of retrieval accuracy, robustness against scale variations, and the impact of both various components and hyper-parameters.

% %*******************************************************
% %*******************************************************
% padding pattern
\begin{figure}[!t]
    \centering
    \includegraphics[width=1.0\linewidth]{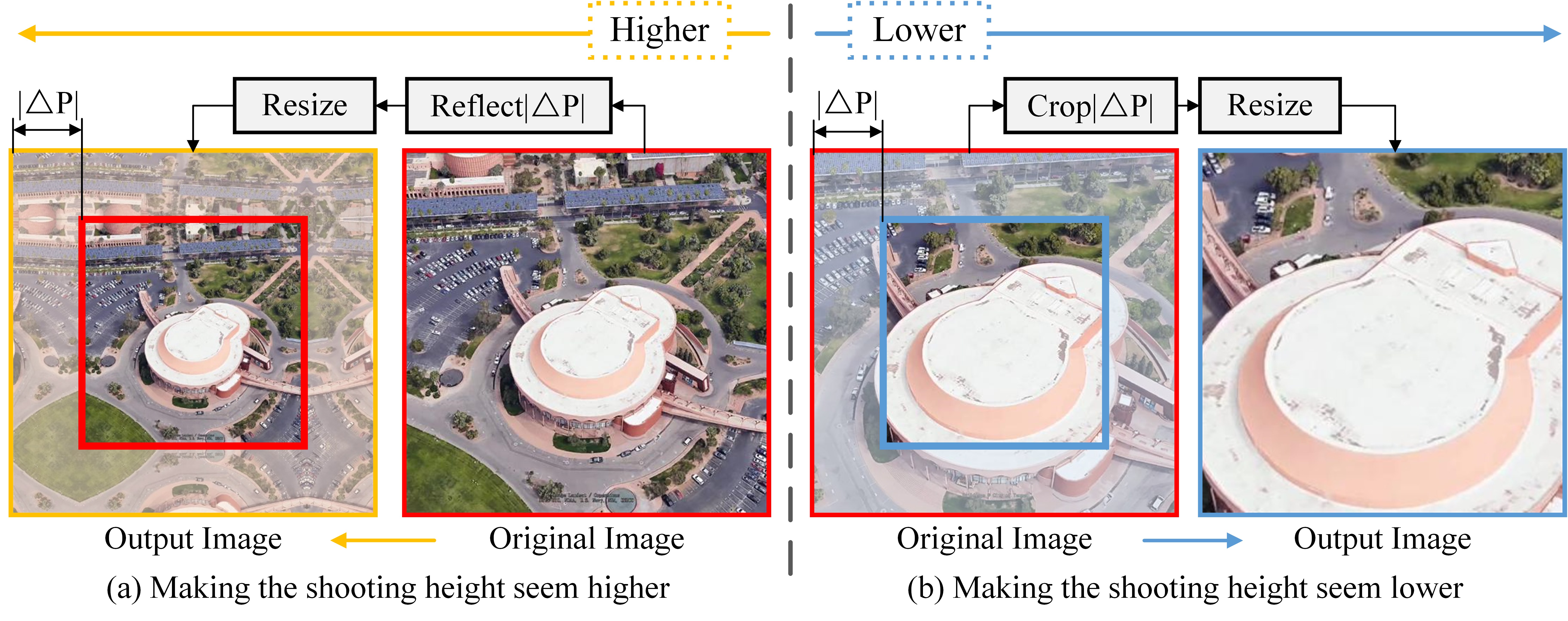}
    \caption{The data augmentation strategy to extend the shooting height range of drone-views from the visual perspective.
    (a) Making the shooting height seem higher: given a drone image, mirror a strip $\Delta$P pixels wide around the input image and resize it to initial resolution;
    (b) Making the shooting height seem lower: given a drone image, crop out a square ring $\Delta$P pixels wide around the input image and resize it to initial resolution.}
    \label{change image distance}
\end{figure}
% %*******************************************************
% ********************************************************

\subsection{Datasets and Evaluation Metrics}\label{Datasets and Evaluation Metrics}
\textbf{Datasets.}
The main experiments are performed based on the University-1652~\cite{zheng2020university} dataset for the following reasons: 1) University-1652 contains diverse scenarios, including 1652 university buildings worldwide, where each building comprises 1 satellite-view image and 54 drone-view images.
The training set includes 701 buildings of 33 universities,
and the other 951 buildings of the remaining 39 universities are classified as the test set.
2) drone-view images exhibits considerable scale variation, with height ranging from 123.5m to 256m~(\textit{i.e.}, $H_{D}\in[123.5,256]$), 
supporting the verification of the anti-scale robustness of SaLPN.
Referring to LPN~\cite{wang2021each}, the test set is equally divided into three parts according to the shooting height of drone-view: \textbf{Long}, \textbf{Middle}, and \textbf{Short}.
Then, comparing satellite-view images with drone-view images, we set the value of $H_{S}$ to 189.75m, which corresponds to the midpoint of the drone height range.

% ********************************************************
% ********************************************************
\begin{table*}[!t]
\centering
\caption{comparison with the state-of-the-art results reported on University-1652. The compared method are divided into two groups, the ResNet50-based methods at the top, and the VIT-based methods at the bottom. Best and second best performance are in red and blue colors.}
\label{comparison with SOTA}
\resizebox{0.95\linewidth}{!}{
\begin{tabular}{ccccccccccccccc}
\hline
\multirow{2}{*}{Methods} & \multirow{2}{*}{Publication} & \multirow{2}{*}{Backbone} &  & \multicolumn{2}{c}{Standard Test set} &  & \multicolumn{2}{c}{Short~($\Delta P$=-100)} &  & \multicolumn{2}{c}{Long~($\Delta P$=100)} &  & \multicolumn{2}{c}{Average} \\ \cline{5-6} \cline{8-9} \cline{11-12} \cline{14-15} 
                        &                              &                           && R@1           & AP           && R@1              & AP              && R@1             & AP             && R@1           & AP          \\ \hline
Instance loss~\cite{zheng2020university}  &  ACMMM'20  &  ResNet-50  &&  60.16  & 64.98   &&  37.62   &  43.36  &&  23.12  &  28.88   &&  40.30  &  45.74      \\
LPN~\cite{wang2021each}  &  TCSVT'21  &   ResNet-50   &&   81.48      &    84.06   &&    47.24  &  52.93   &&  35.57   &  41.51    &&    54.76           &   59.50     \\
CA-HRS~\cite{lu2022content}  &  ACCV'22  &  ResNet-50  && 81.00  &  83.64   &&    48.55   &  54.21  &&    34.94    &   40.94   &&   54.83        &   59.59    \\
MCCG~\cite{shen2023mccg}   &  TCSVT'23  &  ResNet-50  &&  70.94  &  75.04  &&  34.54    &  40.60   &&   25.06   &   31.25   &&   43.51   &  48.96   \\
Sample4Geo~\cite{deuser2023sample4geo}  &   ICCV'23  &   ResNet-50   &  &  78.62  &  82.11  &&  53.30  &  59.05     &&  41.03   &   47.73    &&   57.65     &   62.96    \\
$D^{2}$-GeM~\cite{wang2024rethinking}   &   SPL'24 &  ResNet-50    &&   \Second{84.49}   &  \Second{86.81}    &&   55.55   &    60.92   &&   37.30      &   43.48    &&   59.11   &   63.73    \\
DWDR(w.BT)~\cite{wang2024learning}   &  TGRS'24  &  ResNet-50  &&   71.79  & 75.71  &&   49.16   &   54.75   &&  37.24   &  43.90   &&  52.73    &  58.12      \\
DWDR(w.LPN)~\cite{wang2024learning}   &  TGRS'24  &  ResNet-50  &&   83.18    &   85.63   &&  51.29    &   56.71  &&   36.36  &   42.71  &&  56.94    &  61.68   \\
SDPL~\cite{10587023}   & TCSVT'24  &  ResNet-50  &&   \First{85.19}  &   \First{87.43}  &&  \Second{60.24}   &    \Second{65.10}  &&   \Second{48.29}   &   \Second{54.32}   &&  \Second{64.57} & \Second{68.95}       \\
SaLPN    &     --    &   ResNet-50   &&   84.37  &   86.51  &&    \First{64.89}  &   \First{68.86}   &&     \First{64.08}  &    \First{68.75}   &&   \First{71.11}   &  \First{74.71}   \\
\hdashline
Instance loss~\cite{zheng2020university}   &   ACMMM'20   &    ViT-S  &&  80.34   &   83.05   &&    68.01    &  71.83  &&   61.23   &   65.99  &&  69.86  &     73.62      \\
LPN~\cite{wang2021each}   &  TCSVT'21  &   ViT-S   &&  87.07      &  88.91  &&  69.62  &   73.48  &&   69.81  &  73.98   &&   75.50  &  78.79     \\
FSRA~\cite{dai2021transformer} &  TCSVT'21  &   ViT-S  &&  85.10    &    87.19    &&    71.48   &    75.17    &&  65.67   &  70.04   &&   74.08    &    77.46    \\
SDPL~\cite{10587023}   &  TCSVT'24   &  ViT-S   &&    85.57   &      87.61   &&    68.90   &  72.76  &&   70.56   &   74.65  &&   75.01 &    78.34    \\
Safe-Net~\cite{lin2024self}  &    TIP'24   & ViT-S  &&    \First{88.30}     &     \First{89.93}    &&    \Second{73.32}    &    \Second{76.80}  &&     \Second{73.86} &  \Second{77.51}         &&      \Second{78.49}       &      \Second{81.41}      \\
SaLPN   &  --   &  ViT-S   &&  \Second{88.19}  &  \Second{89.79} &&    \First{73.45}   &  \First{76.91} &&  \First{77.44}   &   \First{80.69}  &&   \First{79.69}   &   \First{82.46}  \\
\hline
\end{tabular}
}
\end{table*}
% ********************************************************
% ********************************************************
% ********************************************************

To further expand the height range of drone-views, we design a data augmentation strategy to simulate more extreme settings.
As shown in Fig.~\ref{change image distance}, we introduce a parameter $\Delta P$ to adjust image scales~(equivalent to drone height), which can be expressed as:
($\Delta P$$>$0) means that making the height of drone-view seem higher~(see Fig.~\ref{change image distance}(a)), while ($\Delta P$$<$0) means that making the height of drone-view seem lower~(see Fig.~\ref{change image distance}(b)).
$|\Delta P|$ denotes the degree of scale adjustment.
Then, we roughly estimate the corresponding new height of synthesized images with $\Delta P$ as:
\begin{equation}\label{eq:height}
    H_{D}' = H_{D} + \lambda_{aug}\cdot \Delta P,
\end{equation}
where $\lambda_{aug}$=0.7 is the empirical value, estimated by the visual scale and height of the existing drone-view images.
By applying the aforementioned strategy to \textbf{Long} and \textbf{Short} sets, we generate test sets with extreme scales, named as \textbf{Long}~($\Delta P$$>$0) and \textbf{Short}~($\Delta P$$<$0), respectively.
Note that drone-views within \textbf{Long}~($\Delta P$$>$0) exhibit higher heights~(meaning smaller image scale), whereas those within \textbf{Short}~($\Delta P$$<$0) have lower heights~(meaning larger image scale).

The SUES-200~\cite{zhu2023sues} dataset collects drone-view images for 200 scenes at four heights~(\textit{i.e.}, 150m, 200m, 250m, 300m) independently, where each scene contains 50 drone-view images and 1 satellite-view image.
The training set includes 120 buildings, and the other 80 buildings are classified as the test set.
The training and testing of models are performed with data of the same height.
We highlight that despite the greater height differences of drone-captured images in SUES-200, SUES-200 exhibits weaker scale differences compared to University-1652 due to the different post-processing methods.

\textbf{Evaluation Metrics.}
To evaluate geo-localization performance, we employ Recall@K~(R@K) and Average Precision~(AP).
R@K refers to the proportion of correctly matched images in the top-K of the ranking list, and AP measures the area under the Precision-Recall curve, considering the ranking of all positive images.

\begin{table}[!t]
\centering
\caption{comparison with the state-of-the-art results reported on SUES-200. The input image size is 256$\times$256, the number of partitions $N$ is 4, and hyper-parameter $\alpha$ is 4.0. Best and second best performance are in red and blue colors.}
\label{comparison with SOTA on SUES-200}
\resizebox{1.0\linewidth}{!}{
\begin{tabular}{lccccccccc}
\hline
\multirow{2}{*}{Methods} &  & \multicolumn{2}{c}{150m} & \multicolumn{2}{c}{200m} & \multicolumn{2}{c}{250m} & \multicolumn{2}{c}{300m} \\
                         && R@1         & AP         & R@1         & AP         & R@1         & AP         & R@1         & AP         \\ \hline
SUES-200~\cite{zhu2023sues}    && 55.65       & 61.92      & 66.78       & 71.55      & 72.00       & 76.43      & 74.05       & 78.26      \\
LCM~\cite{ding2020practical}    && 43.42       & 49.65      & 49.42       & 55.91      & 54.47       & 60.31      & 60.43       & 65.78      \\
LPN~\cite{wang2021each}   && 61.58       & 67.23      & 70.85       & 75.96      & 80.38       & 83.80      & 81.47       & 84.53      \\
FSRA~\cite{dai2021transformer}   &&  \Second{68.25}    & \Second{73.45}  & \Second{83.00}  & \Second{85.99}   & \Second{90.68}    & \Second{92.27}  & \Second{91.95}   & \Second{93.46}  \\
SaLPN   &&       \First{81.92}  &    \First{85.23}     &    \First{87.90}  &  \First{90.02}    &    \First{92.78}  &     \First{94.24}  &   \First{93.03}   &    \First{94.37}     \\ \hline
\end{tabular}
}
\end{table}

\subsection{Implementation Details}\label{Model Settings and Training Details}

\textbf{Model Settings.}
For the SaLPN framework, we set the number of partitions to $N$=4, and partition adjustment factor $\alpha$=14.0 by default.
For SRGS, we adopt Chebyshev distance to generate coordinate map $CM$, and set the heatmap segmentation threshold $\delta$=0.5.
For the University-1652 dataset, we roughly set drone-views height $H_{D}\in[123.5,256]$ and satellite-views height $H_{S}$=$189.75m$, respectively.

\textbf{Training Details.}
By default, we resize each input image to 512$\times$512 for both training and testing phases, and employ the ResNet-50~\cite{he2016deep} to extract visual features.
In training, we set the initial learning rate of 0.0001 for the backbone, and the rest of learnable parameters are set to 0.001.
Stochastic gradient descent~(SGD) is adopted with batch size 4, momentum 0.9, and weight decay 0.0005 for optimization.
The model is trained for 120 epochs, and the learning rate is decayed by 0.1 after 80 epochs.
We also employ random horizontal image flipping as data augmentation.
During testing phase, Euclidean distance is adopted to measure the similarity between the query images and candidate images in the gallery set.
% Experiments are performed on an NVIDIA RTX 3090 GPU with 24 GB of memory.

% %*******************************************************
% 43/55/09
% %*******************************************************
\begin{figure*}[!t]
    \centering
    \includegraphics[width=0.90\linewidth]{img/fig6.jpg}
    \caption{Image retrieval results obtained with SaLPN, LPN and ResNet. Specifically, we present the correspondingly numbered image retrieval outcomes from the original test set~(a), \textbf{Long}~($\Delta P$=100)~(b), and \textbf{Short}~($\Delta P$=-100)~(c), respectively.}
    \label{Image retrieval results}
\end{figure*}
% %*******************************************************
% ********************************************************

\subsection{Comparison with the State-of-the-art}\label{Comparison with the State-of-the-arts}
\textbf{Performance on University-1652.}
As shown in Table~\ref{comparison with SOTA}, we compare SaLPN with other methods on University-1652.
Except for the standard test set provided, we employ Eq.~(\ref{eq:height}) to generate two supplementary test sets, named \textbf{Long}~($\Delta P$=100) and \textbf{Short}~($\Delta P$=-100), which exhibit more extreme height of drone-views.
In other words, \textbf{Long}~($\Delta P$=100) and \textbf{Short}~($\Delta P$=-100) have more significant cross-view scale differences.
The quantitative results are divided into two groups, ResNet-based and ViT-based methods.
On the ResNet-based track, SaLPN achieves 84.37\% R@1 and 86.51\% AP on default test set, 64.89\% R@1 and 68.86\% AP on \textbf{Short}~($\Delta P$=-100) set and 64.08\% R@1, and 68.75\% AP on \textbf{Long}~($\Delta P$=100) set, respectively.
On the default test set, SaLPN achieves competitive performance with recent SDPL~\cite{10587023} and clearly outperforms other solutions, such as $D^2$-GeM~\cite{wang2024rethinking} and Sample4Geo~\cite{deuser2023sample4geo}.
Moreover, it is remarkable that SaLPN yields superior results on both the \textbf{Short}~($\Delta P$=-100) and \textbf{Long}~($\Delta P$=100) compared with partition-based LPN~\cite{dai2021transformer} and SDPL~\cite{10587023}.
Specifically, compared with LPN, SaLPN improves R@1 from 47.24\%, 35.57\% to 64.89\%~(+17.65\%), 64.08\%~(+28.51\%), respectively.
On the ViT-based track, SaLPN is compatible with the ViT framework and has surpassed the sate-of-the-art methods on \textbf{Short}~($\Delta P$=-100) and \textbf{Long}~($\Delta P$=100), achieving an average R@1 and AP of 79.69\% and 82.46\%.
Compared with Instance Loss~\cite{zheng2020university} and FSRA, SaLPN significantly improves retrieval accuracy in all scenarios, with an average R@1 improvement of 9.83\% and 5.61\%.
This shows that SaLPN is robust to scenes with inconsistent cross-view scales.

\textbf{Performance on SUES-200.}
To further validate the effectiveness of SaLPN, we report the results on SUES-200~\cite{zhu2023sues}.
As shown in Table~\ref{comparison with SOTA on SUES-200}, SaLPN achieves competitive results under different height.
Compared with FSRA~\cite{dai2021transformer} at four heights, SaLPN boosts the R@1 from 68.25\%, 83.00\%, 90.68\%, 91.95\% to 81.92\%~(+13.67\%), 87.90\%~(+4.90\%), 92.78\%~(+2.10\%), 93.03\%~(+1.08\%), respectively.
Different from drone-view images with scales variations in University-1652, the training set in SUES-200 has a uniform scale at a specific height.
Besides, we observe that the cross-view scale difference in SUES-200 is weaker than University-1652.
Therefore, we only leverage HAAS in the testing phase and experimentally determine $\alpha$=4.0.

\textbf{Qualitative Results.}
Moreover, we visualize some retrieved results in Fig.~\ref{Image retrieval results}.
To prove the anti-scaling robustness of SaLPN, we provide same numbered results generated by baseline~\cite{he2016deep} and LPN~\cite{wang2021each} on different test sets.
For the original test set, both LPN and SaLPN are able to retrieve the correct satellite labels.
As shown in Fig.~\ref{Image retrieval results}~(b) and~(c), when the drone-view images undergo more drastic scale variations, it can be observed that SaLPN still achieves correct results.
This benefits from the fact that SaLPN explicitly aligns contents of partition pairs, thus mining robust contextual information.
For example, as shown in Group 3 of Fig.~\ref{Image retrieval results}~(b), while the target in the satellite-view is discernible, the target size in the drone-view is notably smaller compared to that in the satellite-view. 
This discrepancy leads to a misalignment of features extracted by LPN with fixed templates, resulting in inaccurate retrieval results.
On the other hand, as shown in Group 2 of Fig.~\ref{Image retrieval results}~(c), drone-view with lower height implies less environmental information and larger scale differences of cross-view images.
Compared to LPN, SaLPN can dynamically adjust partition sizes to boost the alignment of local features, but also refine part-level features, yielding accurate retrieval results.

% **************************************************
% **************************************************
% **************************************************
\begin{table*}[!t]
\centering
\caption{ablation study on robustness of SaLPN to scale-inconsistent scenarios. 
we reported the R@1 accuracy for baseline, LPN, FSRA, SDPL and SaLPN on various test sets. 
The experimental results are divided into two groups: ResNet50-based at the top and ViT-based at the bottom.
best and second best performance are in red and blue colors, respectively.}
\label{table2 robustness}
\resizebox{0.95\linewidth}{!}{
\begin{tabular}{lcccccccccccccccccc}
\hline
\multirow{2}{*}{Test sets} &  & \multicolumn{6}{c}{\textbf{Short}~($\Delta P\leq 0$)}          &  & \textbf{Middle} &  & \multicolumn{6}{c}{\textbf{Long}~($\Delta P\geq 0$)}           &  & \multirow{2}{*}{Average} \\  \cline{3-8} \cline{10-10} \cline{12-17}
        &  & -150 & -100 & -75 & -50 & -25 & 0 && --     && 0 & +25 & +50 & +75 & +100 & +150 &  &                          \\ \hline
Instance loss~\cite{zheng2020university}   &&  9.26  &   37.62   &  48.40   &  56.19   &   61.27  & 64.81   &&    63.58    &&  52.07  &   40.40  &  33.82   &  27.88   &  23.12   &  16.25    &&     41.12  \\
FSRA~\cite{dai2021transformer} &&  7.39    &  33.24   &  46.18   &  54.51   &  59.87   &  64.15  &&   65.07  &&  48.41  &  36.50  &   29.43  &  24.53   &  20.95    &   26.51   && 39.74       \\
LPN~\cite{wang2021each}      &&   9.73   &   47.24   &  64.19   &  74.82   &   80.49  &  84.11  &&   \Second{86.30}     &&  74.01  &  59.38   &  49.58   &  41.20   &   35.57   &  28.31    &&   56.53     \\
SDPL~\cite{10587023}     &&  19.36    &   \Second{60.24}   &   72.30  &  \Second{79.25}   &   \First{83.85}  & \First{86.39}   &&    \First{88.56}    &&  \Second{80.58}  &  69.92  &   62.25  &  54.76   &   48.29   &  39.57    &&   65.02  \\ \hdashline
SaLPN~(w/o HAAS)   &&   \Second{20.56}   &  60.08    &  \Second{72.33}   &  78.89   &   82.52  &  84.75  &&     86.27   &&  79.63   &   \Second{70.77}  &  \Second{64.24}   &  \Second{57.63}   &  \Second{51.13}    &  \Second{39.34}    &&    \Second{65.24}    \\
SaLPN    &&   \First{25.68}   &   \First{64.89}   &   \First{74.95}  &  \First{80.20}   &   \Second{83.54}  & \Second{85.09}   &&   86.16   &&  \First{81.88}  &  \First{75.95}   &  \First{72.24}   &  \First{68.47}   &   \First{64.08}   &  \First{55.48}    &&  \First{70.66}    \\ 
\hline
\hline
Instance loss~\cite{zheng2020university} &&   \First{41.06}  &  68.01  &  74.39   &  77.96   &  79.77  &   80.49 &&   81.87 &&  78.66  &  72.98   &  69.16   &  65.73  &   61.23   &  54.45    &&     69.67    \\
FSRA~\cite{dai2021transformer} &&  \Second{39.35}  &  71.48   &  78.71   &  82.94   &  84.44   & 85.45   &&  86.39  &&  83.44  &   78.86  &  74.73   &  70.72   &  65.67    &   57.26  &&   73.80     \\
LPN~\cite{wang2021each}  &&   32.93   &  69.62    &   78.32  &  83.13   &  85.97   &  87.11  &&   88.41     &&  85.71  &  81.33   &  78.12   &  73.80   &   69.81   &   62.01   &&    75.09                      \\
SDPL~\cite{10587023}     &&  33.93    &   68.90   &   77.22  &  82.07   &  84.62   &  85.45  &&     86.99   &&  84.27  &  80.27   &  77.59   &  74.24   &  70.56    &   63.72   &&    74.60    \\ 
Safe-Net~\cite{lin2024self}     &&  38.96    &  73.32   &   \First{80.85}  &   \First{85.51}  &    \First{87.69}  &  \First{87.91}  &&    \First{89.36}    &&  \Second{87.64}  &  \Second{84.70}   &  \Second{81.68}   &  \Second{78.09}   &    73.86  &   \Second{66.89}   &&   \Second{78.18}    \\ 
\hdashline
SaLPN~(w/o HAAS)  &&   38.79   &  \First{73.66}    &  80.66   &   84.39  &  86.59   &  87.34  &&    89.01    &&  87.46  &  84.05   &   81.13  &   78.01  &   \Second{74.66}   &   66.20   &&   77.84   \\
SaLPN  &&  37.98    &   \Second{73.45}   &  \Second{80.82}   &  \Second{85.01}   &  \Second{87.13}   &  \Second{87.69}   &&    \Second{89.16}    &&  \First{87.73}   &   \First{84.97}  &  \First{82.83}   &  \First{80.48}   & \First{77.44}    &   \First{71.96}  &&    \First{78.97}   \\ 
\hline
\end{tabular}
}
\end{table*}
% *****************************************************
% *****************************************************
% *****************************************************

% %*******************************************************
% %*******************************************************
\begin{figure*}[!t]
    \centering
    \includegraphics[width=1.0\linewidth]{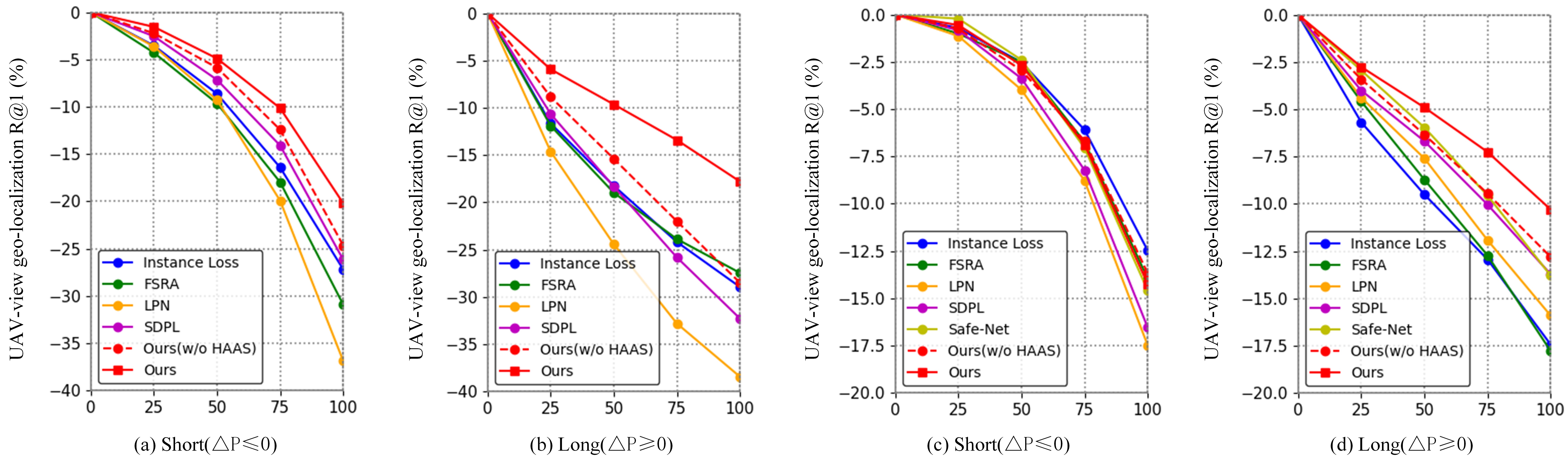}
    \caption{Performance degradation of different methods on \textbf{Short}~($\Delta P\leq 0$) and \textbf{Long}~($\Delta P\geq 0$). Vertical axis: magnitude of decrease in R@1. Horizontal axis: degree of height adjustment $|\Delta P|$. (a) and (b) show the results of ResNet-based methods, while (c) and (d) show the results of ViT-based methods.}
    \label{performance degradation}
\end{figure*}
% %*******************************************************
% ********************************************************

\subsection{Ablation Studies and Further Discussion }\label{Ablation Studies}
To verify the effectiveness of SaLPN, we conduct extensive ablation studies, including: 
\begin{itemize}
    \item To validate the robustness of SaLPN in scale-inconsistent scenarios and its adaptability to diverse backbone networks;
    \item To prove the effectiveness of the proposed square partition strategy, height-aware adjustment strategy and saliency-guided refinement strategy;
    \item To explore different variants of SGRS, including various $CM$, and output features;
    \item To explore the effect of hyper-parameters on model performance, including number of partitions $N$, partition adjustment factor $\alpha$ and input image size.
\end{itemize}

\textbf{Robustness of SaLPN %to 
against scale-inconsistent scenarios.}
As a hard-partition-based method, our method can alleviate the dependence on scale consistency of cross-view images by dynamically adjusting the size of partition pairs.
To fully verify the robustness of SaLPN against scale-inconsistent scenarios, we first synthesize a series of test sets with lower and higher height of drone-views, named \textbf{Short}~($\Delta P$$<$0) and \textbf{Long}~($\Delta P$$>$0), where a larger $|\Delta P|$ represents a larger drone-view scale variation.
Then we compare SaLPN with representative partition-based methods, such as LPN~\cite{wang2021each} and FSRA~\cite{dai2021transformer}.
The results are shown in Table~\ref{table2 robustness}, divided into ResNet50-based track and ViT-based track.
The performance degradation of each method on different datasets is shown in Fig.~\ref{performance degradation}.
On the ResNet50-based track, SaLPN shows the best retrieval accuracy, with significantly less performance degradation than other partition-based methods.
In particular, compared with LPN and SDPL, SaLPN improves average R@1 from 56.53\%, 65.02\% to 70.66\%~(+14.13\%, +5.64\%), respectively.
Furthermore, the integration of HAAS into SaLPN led to a notable improvement in retrieval accuracy in extreme scenarios, with a 5.42\% improvement in the average R@1.
This highlights that HAAS facilitates alignment of fine-grained features from multiple views.
On the ViT-based track, SaLPN also achieves optimal retrieval accuracy across datasets of all scales, with an average R@1 of 78.97\%.
Compared with FSRA and LPN, SaLPN improves average R@1 by 5.11\% and 3.88\%.
This fully validates the robustness of SaLPN to cross-view scale variations, as well as the compatibility with various backbone networks.
Moreover, Fig.~\ref{performance degradation} proves that HAAS can alleviate the performance degradation of SaLPN caused by cross-view scale inconsistency without additional learnable parameters.
It is worth noting that the performance degradation of ViT-based methods is generally lower than that of ResNet-based methods.
We conclude that the ViT framework is capable of capturing long-range feature dependencies, thereby contributing to its robustness against scale changes.
As depicted in Fig.~\ref{performance degradation}~(c), in contrast to ResNet-based methods, the performance degradation of ViT methods is comparable in \textbf{Short}~($\Delta P$$<$0), whereas the baseline exhibits the minimal degradation.
This phenomenon can be attributed to the robust feature extraction capability of the ViT, coupled with the scarcity of effective features in \textbf{Short}~($\Delta P$$<$0) scenarios.

\textbf{Effectiveness of primary components.}
The core components of SaLPN are SPS, HAAS and SGRS.
We construct five derived models to validate the effectiveness of each strategy, and results are shown in Table~\ref{Model structure}.
Firstly, fine-grained partitions generated by SPS contain both local and global features, thus learning rich contextual information.
Compared to the baseline, SPS provides +19.76\% R@1, +21.60\% R@1 and +24.88\% R@1 on three test sets.
When HAAS is combined with SPS, the model's accuracy is notably enhanced in scenarios involving scale inconsistency, while maintaining stable performance in the standard test set.
For instance, after removing HAAS from SaLPN, the R@1 of model~(SPS+SGRS) decreases by 0.83\%, 4.81\%, and 12.95\% on standard test set, \textbf{Short}~($\Delta P$=-100) and \textbf{Long}~($\Delta P$=100), respectively.
We emphasize that drone-views with low height contain incomplete depictions of targets and limited environmental information.
consequently, it is plausible that the performance gain of HAAS is lower in the \textbf{Short} set compared to the \textbf{Long} set.
To further improve the retrieval accuracy of SaLPN, we introduce SGRS to refine salient regions of partitions.
It can be observed that SGRS boosts the retrieval accuracy of SaLPN in various scenarios, yielding 3.68\%, 5.14\% and 3.37\% R@1 improvement over the model~(SPS+HAAS).

% **********************************************************
% **********************************************************
% **********************************************************
\begin{table}[!t]
\centering
\caption{performance comparison of SaLPN framework with different components. $\checkmark$ indicates that the module is applied.}
\label{Model structure}
\resizebox{1.0\linewidth}{!}{
\begin{tabular}{cccccccccccc}
\hline
\multicolumn{3}{c}{Model structure} &  & \multicolumn{2}{c}{Standard Test set} &  & \multicolumn{2}{c}{\textbf{Short}~($\Delta P$=-100)} &  & \multicolumn{2}{c}{\textbf{Long}~($\Delta P$=100)} \\ \cline{1-3} \cline{5-6} \cline{8-9} \cline{11-12} 
SPS       & HAAS      & SGRS       && R@1      & AP      && R@1          & AP         && R@1         & AP     \\ \hline
          &            &            &&     60.16    &   64.98         &&       37.62       &      43.36      &&  23.12           &      28.88      \\
$\checkmark$    &        &     &  &   79.92       &   82.58   &&     59.22  &   63.80   &&   48.00     &    53.64        \\
$\checkmark$         & $\checkmark$          &            &&   80.69          &     83.35       &&       59.75       &   64.17         &&    \Second{60.71}         &    \Second{65.56}        \\
$\checkmark$  &        &    $\checkmark$   &&     \Second{83.54}    &   \Second{85.84}   &&     \Second{60.08}  &  \Second{64.62}   &&  51.13  &  56.98   \\
$\checkmark$   & $\checkmark$   & $\checkmark$   &&    \First{84.37}         &    \First{86.51}        &&    \First{64.89}   &    \First{68.86}    &&     \First{64.08}        &    \First{68.75}        \\ 
\hline
\end{tabular}
}
\end{table}
% **********************************************************
% **********************************************************
% **********************************************************

\begin{figure}[!t]
    \centering
    \includegraphics[width=0.8\linewidth]{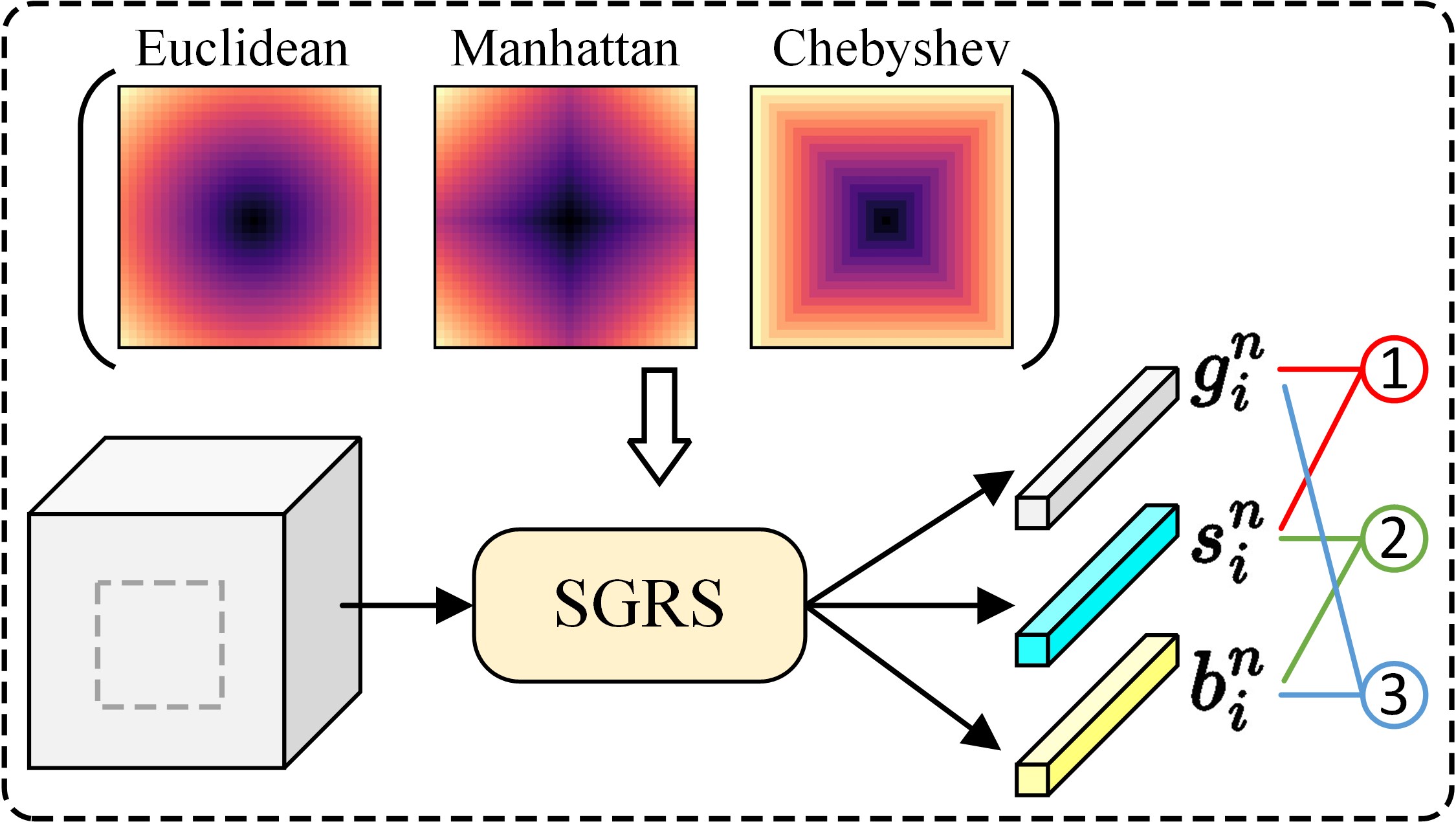}
    \caption{Variants of SGRS in ablation experiments.
    Coordinate map~($CM$) can be calculated by Euclidean distance, Manhattan distance or Chebyshev distance.
    The default SRM output contains global features~($g_{i}^{n}$), salient features~($s_{i}^{n}$), and background features~($b_{i}^{n}$).
    The remaining three output modes can be obtained by combining them in pairs, \textit{i.e.}, \normalsize{\textcircled{\scriptsize{1}}}$g_{i}^{n}+s_{i}^{n}$, \normalsize{\textcircled{\scriptsize{2}}}$s_{i}^{n}+b_{i}^{n}$, \normalsize{\textcircled{\scriptsize{3}}}$g_{i}^{n}+b_{i}^{n}$.}
    \label{different_SRM}
\end{figure}

\textbf{Variants of SGRS.}
As mentioned above, SGRS segments part-level features based on heatmaps to obtain global features, salient features and background features, which are matched independently. 
To fully verify the effectiveness of SGRS, we construct a variety of SGRS variants, including different output combinations as well as coordinate map $CM$, as shown in Fig.~\ref{different_SRM}.
Table~\ref{Variants of SGRS} shows the experimental results.
Regardless of the specific combination of output modes employed by SGRS to refine partitions, it consistently outperforms the model without SGRS~(\textit{i.e.}, only $g_{i}$).
On the other hand, compared to Euclidean distance and Manhattan distance, the utilization of Chebyshev distance within SGRS yields an optimal gain, surpassing that of Manhattan distance by a notable margin.
This phenomenon can be attributed to the fact that $CM$ generated by the Chebyshev paradigm exhibits a structural similarity to the square partition strategy employed.
Therefore, it is within the realm of expectation that the model based on Euclidean distance yields comparable results.
Since the feature segmentation process of SGRS is unsupervised, the injected $CM$ can assist SGRS in judging the importance of features and classifying neighborhood features into the same class.
After removing of $CM$ from SGRS, the average R@1 and AP metrics of SaLPN decreased by 6.48\% and 5.83\%.

% %*******************************************************
% %*******************************************************
% %*******************************************************
% ********************************************************

% **********************************************************
% **********************************************************
% **********************************************************
\begin{table*}[!t]
\centering
\caption{ablation study on different variants of SGRS. 
The experiment aims to explore the impact of different feature combinations and coordinate maps on performance.}
\label{Variants of SGRS}
\resizebox{0.95\linewidth}{!}{
\begin{tabular}{ccccccccccccccccccc}
\hline
\multicolumn{3}{c}{Output features} &  & \multicolumn{3}{c}{Coordinate maps} &  & \multicolumn{2}{c}{Standard Test set} &  & \multicolumn{2}{c}{\textbf{Short}~($\Delta P$=-100)} &  & \multicolumn{2}{c}{\textbf{Long}~($\Delta P$=100)} &  & \multicolumn{2}{c}{Average} \\ \cline{1-3} \cline{5-7} \cline{9-10} \cline{12-13} \cline{15-16} \cline{18-19} 
$g_{i}$         & $s_{i}$         & $b_{i}$        && Euclidean  & Manhattan  & Chebyshev && R@1           & AP           && R@1          & AP         && R@1         & AP         && R@1           & AP          \\ \hline 
$\checkmark$    &            &           &&            &            &           &&    80.69   &    83.35   &&    59.75    &  64.17     &&   60.71   &   65.56   &&    67.05     & 71.02    \\
\hdashline
$\checkmark$  & $\checkmark$  &      &&        &         & $\checkmark$  &&   82.15  &   84.57   &&   \Second{64.36}   &   68.36    &  &  62.00   &   66.72   &  &   69.50  &  73.21    \\
$\checkmark$  &     & $\checkmark$  &&     &       &     $\checkmark$      &&    83.56  &   85.88   &&   62.37  &   66.70  &  &    61.55  &    66.36  &&   69.16   &   72.98  \\
& $\checkmark$   & $\checkmark$   &&          &            &   $\checkmark$  &&  83.99   &   86.14   &&   63.57    &   67.82   &  &   61.49   &  66.19     &&   69.68  &  73.38     \\
\hdashline
$\checkmark$    & $\checkmark$  & $\checkmark$    &&            &            &           &&    79.66   &    82.42  &&   56.38  &  61.51  &&      57.87   &    62.72  &&   64.63    &   68.88      \\
$\checkmark$   & $\checkmark$  & $\checkmark$   && $\checkmark$    &     &     &&    \Second{84.02}  &   \Second{86.20} &&     64.16   &  \Second{68.40}   &  &   \Second{63.16}    &   \Second{67.69}  &&    \Second{70.44}    &  \Second{74.09}      \\
$\checkmark$   & $\checkmark$  & $\checkmark$  &&            & $\checkmark$       &          &&    82.69     &   85.05    &&   59.15  &    63.73   &&  60.76  &  65.71  &&    67.53   &   71.49   \\
\hdashline
$\checkmark$  & $\checkmark$    & $\checkmark$ &&            &            & $\checkmark$      &&  \First{84.37}    &   \First{86.51}       &&    \First{64.89}   &    \First{68.86}  &&  \First{64.08}    &    \First{68.75}   &&   \First{71.11}   &  \First{74.71}    \\
\hline
\end{tabular}
}
\end{table*}
% **********************************************************
% **********************************************************
% **********************************************************

\textbf{Partition adjustment factor $\alpha$.}
Referring to Eq.~(\ref{eq4}), HAAS can adjust the degree of variation in partition size by modifying $\alpha$.
An improperly choice for $\alpha$ may result in a misalignment of content between cross-view partition pairs.
To ensure that $\alpha$ satisfies all test scenarios, $H_D^{max}$ and $H_D^{min}$ in Eq.~\ref{eq6} and \ref{eq7} are from \textbf{Long}~($\Delta P$=+150) and \textbf{Short}~($\Delta P$=-150), respectively.
Then, we set a series of $\alpha$ for experiments, and the results are shown in Table~\ref{different alpha}.
It can be observed that when $\alpha$ is small, the size of partition pairs of SaLPN is almost of the same size, yielding favorable outcomes in the \textbf{Middle} set but suffering from significant performance degradation in extreme scenarios.
When $\alpha$ increases from 0.0 to 14.0, the retrieval accuracy of SaLPN continues to rise in the \textbf{Short}~($\Delta P\leq 0$) and \textbf{Long}~($\Delta P\geq 0$) sets.
In particular, the R@1 of our method improves from 17.84\% and 34.93\% to 25.68\%(+7.84\%) and 55.48\%(+20.55\%) on \textbf{Short}~($\Delta P$=-150) set and \textbf{Long}~($\Delta P$=+150) set, respectively.
The average R@1 of SaLPN also increases from 63.83\% to 70.66\%(+6.83\%).
This verifies that HAAS can enhance the robustness of our model against scale changes.
As $\alpha$ continues to increase, the retrieval accuracy of SaLPN exhibits a slight decline across various \textbf{Short}~($\Delta P\leq 0$) scenes.
Considering the average performance of the model, we set $\alpha$=14.0 as the default setting.

% **********************************************************
% **********************************************************
% **********************************************************
\begin{table*}[!t]
\centering
\caption{performance comparison of SaLPN with various partition adjustment factor $\alpha$.
we reported the R@1 accuracy on three types of datasets, including \textbf{Short}~($\Delta P\leq 0$) set, \textbf{Middle} set and \textbf{Long}~($\Delta P\geq 0$) set.}
\label{different alpha}
\resizebox{0.95\linewidth}{!}{
\begin{tabular}{lcccccccccccccccccc}
\hline
\multirow{2}{*}{Test sets} &  & \multicolumn{6}{c}{\textbf{Short}($\Delta P\leq 0$)}          &  & \textbf{Middle} &  & \multicolumn{6}{c}{\textbf{Long}($\Delta P\geq 0$)}           &  & \multirow{2}{*}{Average} \\  \cline{3-8} \cline{10-10} \cline{12-17}
        &  & -150 & -100 & -75 & -50 & -25 & 0 &  & --     &  & 0 & +25 & +50 & +75 & +100 & +150 &  &                          \\ \hline
$\alpha$=0.0  &&   17.84   &  58.84    &  71.64   &  78.86   & 83.12    &  85.31  &&   86.36     &&  78.79  &   69.52  &  62.24   &   54.71  &  47.69    &   34.93   &&    63.83     \\
$\alpha$=2.0     &&   21.55   &  62.64    &  73.37   & 79.00    &  83.02   &  85.20  &&   \Second{86.73}     &&  79.46  &  69.87   &   63.84  &   57.98  &   52.21   &   40.51   &&     65.79      \\
$\alpha$=4.0     &&  21.71    &  61.99    &  73.60   &  79.78   & 83.22    &  85.29  &&   \First{86.83}     &&  79.93  &  71.70  &  65.67  &  59.70  &  53.61    &   42.04   &&    66.54      \\
$\alpha$=6.0      &&   23.69   &  63.57    &   74.08  &   \First{80.36}  & \First{83.98}   &  \First{85.44}  &&    86.31    && 80.11   &  72.49   &  67.29   &   62.14  &   56.59   &  46.20    &&   67.86      \\
$\alpha$=8.0     &&   23.24 &   63.36   &   73.84  &  80.05   &  83.37   & \Second{85.36}   &&    86.54    &&  80.35  &  72.71  &  68.38  & 63.38   &  58.02    &   47.59   &&   68.17       \\
$\alpha$=10.0     &&   25.66   &  64.66    &  \Second{74.67}   &  \Second{80.28}   &  83.44   &  85.10  &&    86.13    &&  80.61  &   73.46  &   68.99  &   64.02  &   58.82   &    50.20  &&     68.92        \\
$\alpha$=12.0     &&   23.52   &  63.40    &  73.88   &  79.68   &  82.98   &  85.28  &&    86.31    &&  \Second{81.18}  &   \Second{74.85}  &   71.18  &   67.08  &   62.79   &    54.68  &&     69.75        \\
$\alpha$=14.0     &&   \Second{25.68}   &   \First{64.89}   &  \First{74.95}   &  80.20   &   \Second{83.54}  & 85.09   &&    86.16    &&  \First{81.88}  &   \First{75.95} &  \First{72.24}  & \First{68.47}   &  \Second{64.08}    &   \Second{55.48}   &&     \First{70.66}     \\
$\alpha$=16.0     &&  \First{26.12}   &  \Second{64.73}    &  74.45   &  79.74   &  82.63   &  84.27  &&     84.87   &&  80.67  &  74.73   &  \Second{71.35}   &  \Second{67.83}   &  \First{64.42}    &  \First{56.77}    && \Second{70.19}    \\
\hline
\end{tabular}
}
\end{table*}
% **********************************************************
% **********************************************************
% **********************************************************

\begin{figure}[!t]
    \centering
    \includegraphics[width=0.97\linewidth]{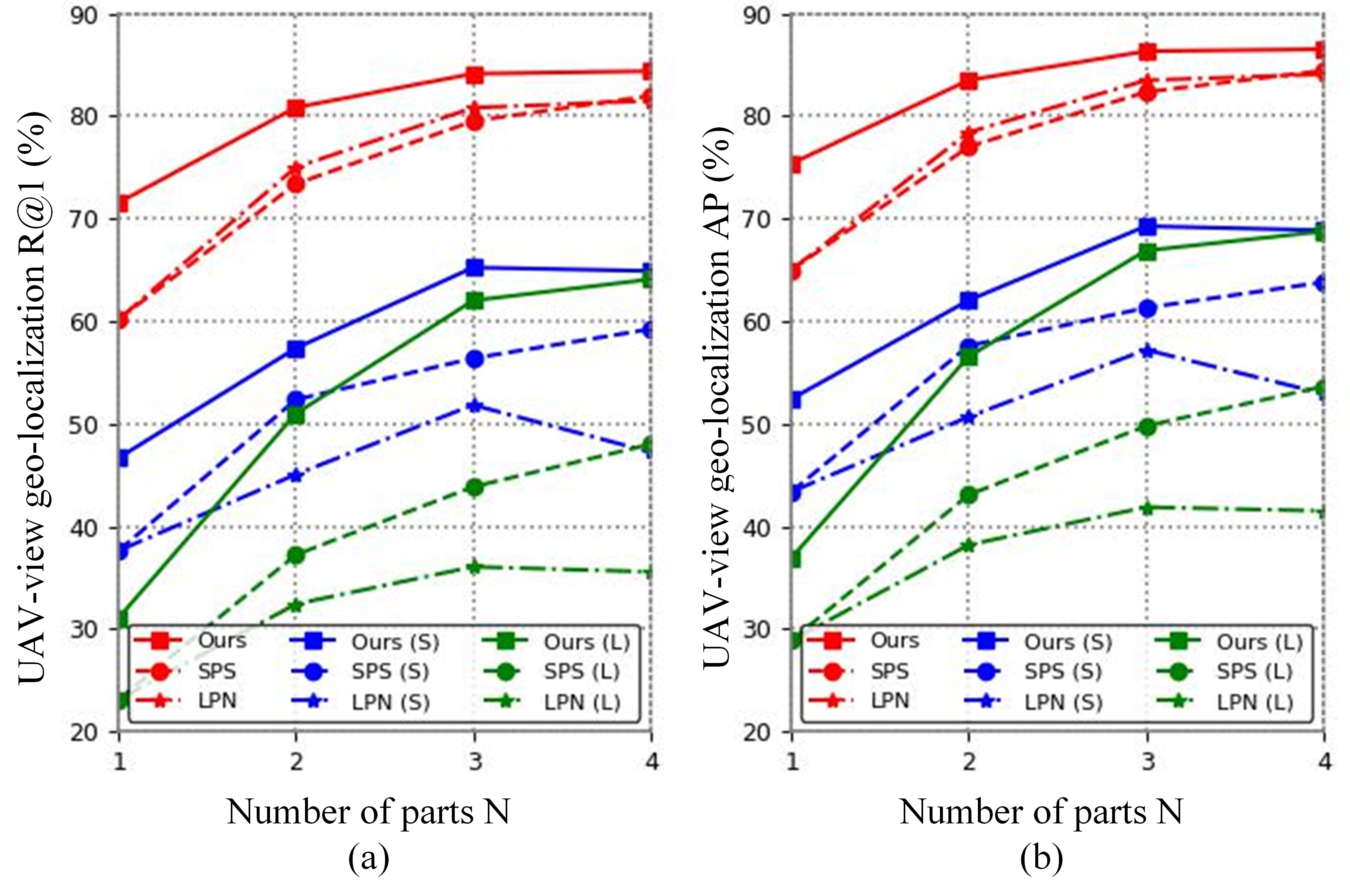}
    \caption{Comparison of LPN, SPS and SaLPN with different numbers of parts. The experiments were conducted on the standard test set, \textbf{Short}~($\Delta P$=-100) and \textbf{Long}~($\Delta P$=100).}
    \label{comparison NSPS}
\end{figure}

\textbf{Number of partitions.}
The number of parts $N$ is a key hyper-parameter in SaLPN.
As shown in Table~\ref{Number of partitions}, we report the quantitative results of SaLPN with different $N$ on the three test-sets.
As the number of parts increases, the retrieval accuracy of SaLPN exhibits a gradual enhancement, particularly in scenarios involving inconsistent scales.
For example, when $N$ increases from 1 to 3, SaLPN improves R@1 by 12.58\%, 18.57\% and 30.96\% in the standard test set, \textbf{Short}~($\Delta P$=-100) and \textbf{Long}~($\Delta P$=-100), respectively.
Further increasing $N$, the retrieval performance of SaLPN tends to be stable with a slight improvement, because 3 or 4 partitions are enough to exploit rich contextual information.
As the basic component of SaLPN, SPS divides high-level features into fixed-size fine-grained partitions.
To thoroughly validates the anti-scaling robustness of SPS, we also report results for LPN~\cite{wang2021each} and SPS under various number of partitions, as shown in Fig.~\ref{comparison NSPS}.
It is evident that SPS exhibits comparable performance to LPN on the standard test set, while outperforming LPN significantly in scale-inconsistent scenarios.
When SPS is combined with HAAS and SGRS, model~(\textit{i.e.}, SaLPN) with various $N$ achieves significant performance improvement in each test set.

% **********************************************************
% **********************************************************
% **********************************************************
\begin{table}[!t]
\centering
\caption{performance comparison of SaLPN with different number of parts. $N$ denotes the number of parts generated by $SPS$.}
\label{Number of partitions}
\resizebox{1.0\linewidth}{!}{
\begin{tabular}{cccccccccc}
\hline
Numbers &  & \multicolumn{2}{c}{Standard Test set} &  & \multicolumn{2}{c}{\textbf{Short}~($\Delta P$=-100)} &  & \multicolumn{2}{c}{\textbf{Long}~($\Delta P$=100)}  \\ \cline{3-4} \cline{6-7} \cline{9-10}  
$N$  && R@1      & AP       && R@1       & AP       && R@1        & AP      \\ \hline
1       &&     71.54      &     75.35       &&    46.66          &     52.41      &&      31.05    &     36.90       \\
2     &&      80.83   &    83.49     &&   57.36  &   62.03         &&     50.99      &     56.56                 \\
3      &&      \Second{84.12}         &    \Second{86.31}          &&     \First{65.23}         &     \First{69.26}       &&     \Second{62.01}     &     \Second{66.88}        \\
4    &&     \First{84.37}  &  \First{86.51}    &&     \Second{64.89}    &    \Second{68.86}      &  &      \First{64.08}     &      \First{68.75}                \\ 
\hline
\end{tabular}
}
\end{table}
% **********************************************************
% **********************************************************
% **********************************************************

% %*******************************************************
% %*******************************************************
% %*******************************************************
% ********************************************************

\begin{figure}[!t]
    \centering
    \includegraphics[width=1.0\linewidth]{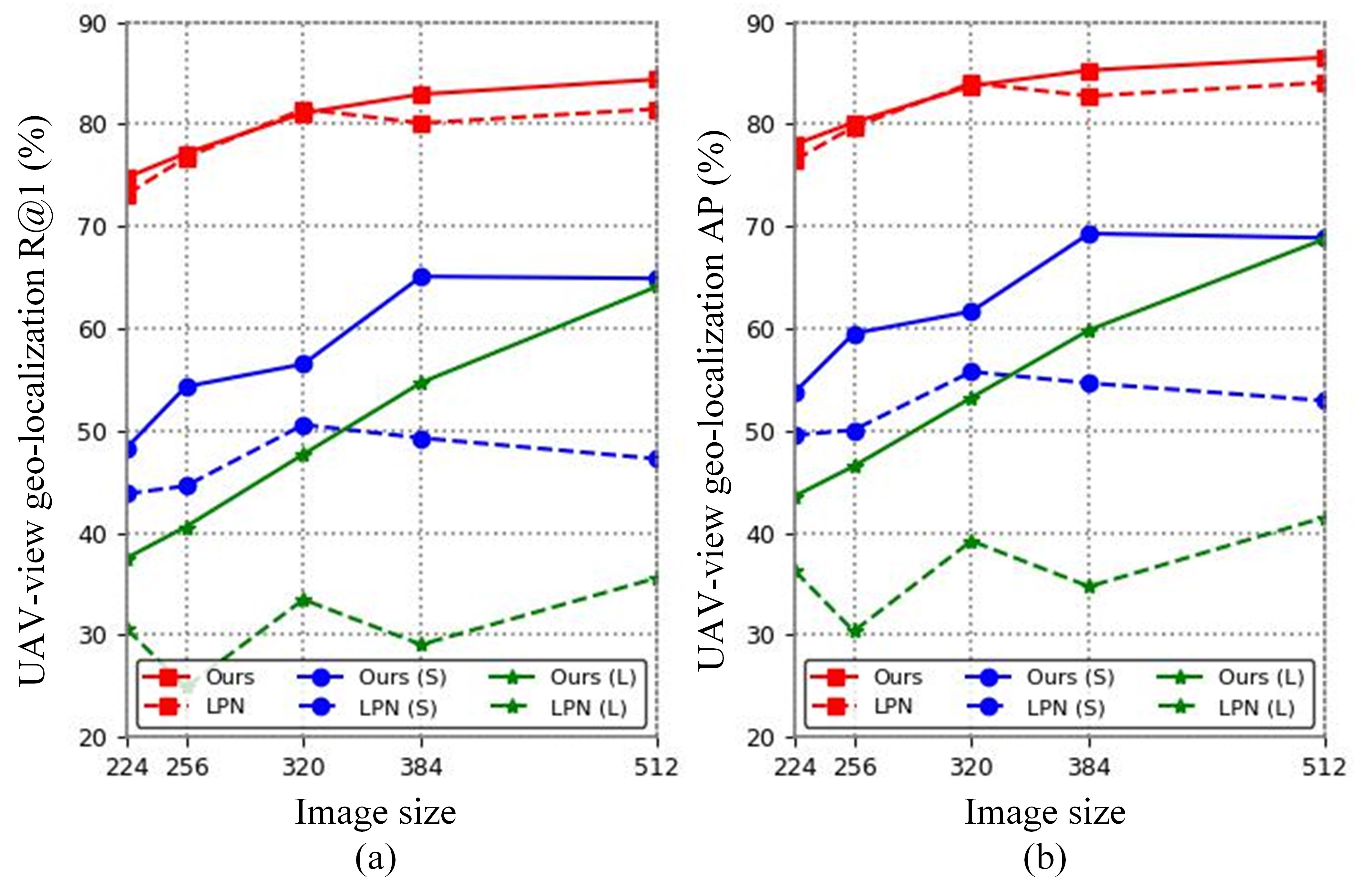}
    \caption{Impact of different input sizes on R@1 and AP. The experiments were conducted on the standard test set, \textbf{Short}~($\Delta P$=-100) and \textbf{Long}~($\Delta P$=100).}
    \label{comparison imagesize}
\end{figure}

\textbf{Input image size.}
A small training size compresses the fine-grained information of the input image, thus compromising the discriminative representation learning.
We select five input image resolutions to investigate the impact of input image size on model performance.
To guarantee that the partition resolution remains positive, we adjust the partition adjustment factor $\alpha$ in accordance with variations in the input image resolution.
As shown in Table~\ref{input image size}, when the input image resolution changes from 224 to 512, model performance improves continuously on multiple datasets.
Furthermore, with the image resolution increasing from 320 to 512, the gain of SaLPN is 3.29\% on the standard test set, while the improvement is 8.38\% and 16.40\% on \textbf{Short}~($\Delta P$=-100) and \textbf{Long}~($\Delta P$=100), respectively.
We also present the results of LPN~\cite{wang2021each}, the most relevant study, under different input picture sizes, as depicted in Fig.~\ref{comparison imagesize}.
With the increase of image resolution, the accuracy of LPN and SaLPN is equivalent in the standard test set, but the performance gap between them gradually widens in extreme scenes.
It can also be found that the performance of LPN is unstable in \textbf{Short}($\Delta P$=-100) and \textbf{Long}($\Delta P$=100).
This indicates that SaLPN can extract the rich fine-grained information contained in high-resolution images.

% **********************************************************
% **********************************************************
% **********************************************************
\begin{table}[!t]
\centering
\caption{performance comparison of SaLPN with different input image size.}
\label{input image size}
\resizebox{1.0\linewidth}{!}{
\begin{tabular}{cccccccccc}
\hline
\multirow{2}{*}{Image size} &  & \multicolumn{2}{c}{Standard Test set} &  & \multicolumn{2}{c}{\textbf{Short}~($\Delta P$=-100)} &  & \multicolumn{2}{c}{\textbf{Long}~($\Delta P$=100)}  \\ \cline{3-4} \cline{6-7} \cline{9-10}  
  && R@1           & AP           && R@1          & AP         && R@1         & AP         \\ \hline
224$\times$224    &&     74.79    &    78.01    &&      48.32        &     53.75   &&     37.53      &      43.59      \\
256$\times$256    &&   77.14     &  80.16    &&     54.30         &        59.50    &&   40.58     &  46.51     \\
320$\times$320    &&   81.08     &    83.74      &&   56.51           &     61.66       &&     47.68    &   53.22     \\
384$\times$384    &&  \Second{82.92}    &   \Second{85.27}       &&   \First{65.08}   &     \First{69.28} &&   \Second{54.69}     &   \Second{59.84}    \\ 
512$\times$512    &&     \First{84.37}  &  \First{86.51}    &&     \Second{64.89}    &    \Second{68.86}      &&      \First{64.08}     &      \First{68.75}    \\ 
\hline
\end{tabular}
}
\end{table}
% **********************************************************
% **********************************************************
% **********************************************************

% %*******************************************************
% %*******************************************************
% %*******************************************************
% ********************************************************

%%%%%%%%%%%%%%%%%%%%%%%%%%%%%%%%%%%%%%%%%%
\section{Conclusion}\label{Conclusions}
In this paper, we study the UAV-view geo-localization in scale-inconsistent scenarios, and propose a simple and effective part-based representation learning framework called SaLPN.
Specifically, we introduce a square partition strategy to capture fine-grained features while retaining the global structure.
Adhering to the rule that objects in drone-views appear larger when closer and smaller when farther, we further introduce a height-aware adjustment strategy.
Based on the height ratio of cross-view images, we dynamically adjust the partition size of the drone branch to promote the content consistency of partition pairs, thereby achieving accurate cross-view matching against scale variations.
Moreover, we explore a saliency-guided refinement strategy to refine part-level features into salient features and background features, further improving cross-view image matching performance without compromising the robustness of our model to scale variations.
Extensive experiments on University-1652 and SUES-200 validate that the effectiveness and robustness of our method in various scale-inconsistent scenarios.
Ablation experiments show that SaLPN reveals satisfactory compatibility with various backbone networks.
In the future, we plan to collect a real-world dataset containing drone pose information to drive real-world cross-view matching, such as drone-view geo-localization and drone self-positioning.

% \section*{Acknowledgments}
% This should be a simple paragraph before the References to thank those individuals and institutions who have supported your work on this article.

% \section{References Section}
% You can use a bibliography generated by BibTeX as a .bbl file.
%  BibTeX documentation can be easily obtained at:
%  http://mirror.ctan.org/biblio/bibtex/contrib/doc/
%  The IEEEtran BibTeX style support page is:
%  http://www.michaelshell.org/tex/ieeetran/bibtex/
 
% \bibliography{ref}

% You can manually copy in the resultant .bbl file and set second argument of $\backslash${\tt{begin}} to the number of references
%  (used to reserve space for the reference number labels box).

\bibliographystyle{IEEEtran}
\bibliography{ref.bib}

\begin{thebibliography}{10}
\providecommand{\url}[1]{#1}
\csname url@samestyle\endcsname
\providecommand{\newblock}{\relax}
\providecommand{\bibinfo}[2]{#2}
\providecommand{\BIBentrySTDinterwordspacing}{\spaceskip=0pt\relax}
\providecommand{\BIBentryALTinterwordstretchfactor}{4}
\providecommand{\BIBentryALTinterwordspacing}{\spaceskip=\fontdimen2\font plus
\BIBentryALTinterwordstretchfactor\fontdimen3\font minus \fontdimen4\font\relax}
\providecommand{\BIBforeignlanguage}[2]{{%
\expandafter\ifx\csname l@#1\endcsname\relax
\typeout{** WARNING: IEEEtran.bst: No hyphenation pattern has been}%
\typeout{** loaded for the language `#1'. Using the pattern for}%
\typeout{** the default language instead.}%
\else
\language=\csname l@#1\endcsname
\fi
#2}}
\providecommand{\BIBdecl}{\relax}
\BIBdecl

\bibitem{goodrich2023placement}
P.~Goodrich, O.~Betancourt, A.~C. Arias, and T.~Zohdi, ``Placement and drone flight path mapping of agricultural soil sensors using machine learning,'' \emph{Computers and Electronics in Agriculture}, vol. 205, p. 107591, 2023.

\bibitem{khan2017uav}
M.~A. Khan, W.~Ectors, T.~Bellemans, D.~Janssens, and G.~Wets, ``Uav-based traffic analysis: A universal guiding framework based on literature survey,'' \emph{Transportation research procedia}, vol.~22, pp. 541--550, 2017.

\bibitem{zhao2023ms}
L.~Zhao and M.~Zhu, ``Ms-yolov7: Yolov7 based on multi-scale for object detection on uav aerial photography,'' \emph{Drones}, vol.~7, no.~3, p. 188, 2023.

\bibitem{zimmermann2017precise}
F.~Zimmermann, C.~Eling, L.~Klingbeil, and H.~Kuhlmann, ``Precise positioning of uavs--dealing with challenging rtk-gps measurement conditions during automated uav flights,'' \emph{ISPRS Annals of the Photogrammetry, Remote Sensing and Spatial Information Sciences}, vol.~4, pp. 95--102, 2017.

\bibitem{lin2013cross}
T.-Y. Lin, S.~Belongie, and J.~Hays, ``Cross-view image geolocalization,'' in \emph{Proceedings of the IEEE Conference on Computer Vision and Pattern Recognition}, 2013, pp. 891--898.

\bibitem{bansal2011geo}
M.~Bansal, H.~S. Sawhney, H.~Cheng, and K.~Daniilidis, ``Geo-localization of street views with aerial image databases,'' in \emph{Proceedings of the 19th ACM international conference on Multimedia}, 2011, pp. 1125--1128.

\bibitem{castaldo2015semantic}
F.~Castaldo, A.~Zamir, R.~Angst, F.~Palmieri, and S.~Savarese, ``Semantic cross-view matching,'' in \emph{Proceedings of the IEEE International Conference on Computer Vision Workshops}, 2015, pp. 9--17.

\bibitem{zheng2020university}
Z.~Zheng, Y.~Wei, and Y.~Yang, ``University-1652: A multi-view multi-source benchmark for drone-based geo-localization,'' in \emph{Proceedings of the 28th ACM international conference on Multimedia}, 2020, pp. 1395--1403.

\bibitem{shen2023mccg}
T.~Shen, Y.~Wei, L.~Kang, S.~Wan, and Y.-H. Yang, ``Mccg: A convnext-based multiple-classifier method for cross-view geo-localization,'' \emph{IEEE Transactions on Circuits and Systems for Video Technology}, 2023.

\bibitem{lin2022joint}
J.~Lin, Z.~Zheng, Z.~Zhong, Z.~Luo, S.~Li, Y.~Yang, and N.~Sebe, ``Joint representation learning and keypoint detection for cross-view geo-localization,'' \emph{IEEE Transactions on Image Processing}, vol.~31, pp. 3780--3792, 2022.

\bibitem{wang2024learning}
T.~Wang, Z.~Zheng, Z.~Zhu, Y.~Sun, C.~Yan, and Y.~Yang, ``Learning cross-view geo-localization embeddings via dynamic weighted decorrelation regularization,'' \emph{IEEE Transactions on Geoscience and Remote Sensing}, vol.~62, pp. 1--12, 2024.

\bibitem{wang2024rethinking}
T.~Wang, Z.~Yang, Q.~Chen, Y.~Sun, and C.~Yan, ``Rethinking pooling for multi-granularity features in aerial-view geo-localization,'' \emph{IEEE Signal Processing Letters}, 2024.

\bibitem{zheng2023exif}
C.~Zheng, A.~Shrivastava, and A.~Owens, ``Exif as language: Learning cross-modal associations between images and camera metadata,'' in \emph{Proceedings of the IEEE/CVF Conference on Computer Vision and Pattern Recognition}, 2023, pp. 6945--6956.

\bibitem{zheng2024iterated}
C.~Zheng, J.~Zhang, A.~Kembhavi, and R.~Krishna, ``Iterated learning improves compositionality in large vision-language models,'' in \emph{Proceedings of the IEEE/CVF Conference on Computer Vision and Pattern Recognition}, 2024, pp. 13\,785--13\,795.

\bibitem{MCGF}
T.~Feng, Q.~Li, X.~Wang, M.~Wang, G.~Li, and W.~Zhu, ``Multi-weather cross-view geo-localization using denoising diffusion models,'' in \emph{Proceedings of the 2nd Workshop on UAVs in Multimedia: Capturing the World from a New Perspective}, 2024, p. 35–39.

\bibitem{chen2024multi}
Z.~Chen, Z.-X. Yang, and H.-J. Rong, ``Multi-level embedding and alignment network with consistency and invariance learning for cross-view geo-localization,'' \emph{arXiv preprint arXiv:2412.14819}, 2024.

\bibitem{wang2024multiple}
T.~Wang, Z.~Zheng, Y.~Sun, C.~Yan, Y.~Yang, and T.-S. Chua, ``Multiple-environment self-adaptive network for aerial-view geo-localization,'' \emph{Pattern Recognition}, vol. 152, p. 110363, 2024.

\bibitem{10644040}
Q.~Wu, Y.~Wan, Z.~Zheng, Y.~Zhang, G.~Wang, and Z.~Zhao, ``Camp: A cross-view geo-localization method using contrastive attributes mining and position-aware partitioning,'' \emph{IEEE Transactions on Geoscience and Remote Sensing}, vol.~62, pp. 1--14, 2024.

\bibitem{10636268}
P.~Xia, Y.~Wan, Z.~Zheng, Y.~Zhang, and J.~Deng, ``Enhancing cross-view geo-localization with domain alignment and scene consistency,'' \emph{IEEE Transactions on Circuits and Systems for Video Technology}, vol.~34, no.~12, pp. 13\,271--13\,281, 2024.

\bibitem{dai2021transformer}
M.~Dai, J.~Hu, J.~Zhuang, and E.~Zheng, ``A transformer-based feature segmentation and region alignment method for uav-view geo-localization,'' \emph{IEEE Transactions on Circuits and Systems for Video Technology}, vol.~32, no.~7, pp. 4376--4389, 2021.

\bibitem{zhao2024transfg}
H.~Zhao, K.~Ren, T.~Yue, C.~Zhang, and S.~Yuan, ``Transfg: A cross-view geo-localization of satellite and uavs imagery pipeline using transformer-based feature aggregation and gradient guidance,'' \emph{IEEE Transactions on Geoscience and Remote Sensing}, 2024.

\bibitem{li2024geoformer}
Q.~Li, X.~Yang, J.~Fan, R.~Lu, B.~Tang, S.~Wang, and S.~Su, ``Geoformer: An effective transformer-based siamese network for uav geo-localization,'' \emph{IEEE Journal of Selected Topics in Applied Earth Observations and Remote Sensing}, 2024.

\bibitem{li2023transformer}
S.~Li, C.~Liu, H.~Qiu, and Z.~Li, ``A transformer-based adaptive semantic aggregation method for uav visual geo-localization,'' in \emph{Chinese Conference on Pattern Recognition and Computer Vision (PRCV)}.\hskip 1em plus 0.5em minus 0.4em\relax Springer, 2023, pp. 465--477.

\bibitem{zhu2022transgeo}
S.~Zhu, M.~Shah, and C.~Chen, ``Transgeo: Transformer is all you need for cross-view image geo-localization,'' in \emph{Proceedings of the IEEE/CVF Conference on Computer Vision and Pattern Recognition}, 2022, pp. 1162--1171.

\bibitem{wang2021each}
T.~Wang, Z.~Zheng, C.~Yan, J.~Zhang, Y.~Sun, B.~Zheng, and Y.~Yang, ``Each part matters: Local patterns facilitate cross-view geo-localization,'' \emph{IEEE Transactions on Circuits and Systems for Video Technology}, vol.~32, no.~2, pp. 867--879, 2022.

\bibitem{li2024aerial}
H.~Li, T.~Wang, Q.~Chen, Q.~Zhao, S.~Jiang, C.~Yan, and B.~Zheng, ``Aerial-view geo-localization based on multi-layer local pattern cross-attention network,'' \emph{Applied Intelligence}, vol.~54, no.~21, pp. 11\,034--11\,053, 2024.

\bibitem{ge2024multi}
F.~Ge, Y.~Zhang, L.~Wang, W.~Liu, Y.~Liu, S.~Coleman, and D.~Kerr, ``Multi-level feedback joint representation learning network based on adaptive area elimination for cross-view geo-localization,'' \emph{IEEE Transactions on Geoscience and Remote Sensing}, 2024.

\bibitem{ge2024multibranch}
F.~Ge, Y.~Zhang, Y.~Liu, G.~Wang, S.~Coleman, D.~Kerr, and L.~Wang, ``Multibranch joint representation learning based on information fusion strategy for cross-view geo-localization,'' \emph{IEEE Transactions on Geoscience and Remote Sensing}, vol.~62, pp. 1--16, 2024.

\bibitem{nanhua2024mmhca}
C.~Nanhua, T.-s. LOU, and Z.~Liangyu, ``Mmhca: Multi-feature representations based on multi-scale hierarchical contextual aggregation for uav-view geo-localization,'' \emph{Chinese Journal of Aeronautics}, 2024.

\bibitem{lin2024self}
J.~Lin, Z.~Luo, D.~Lin, S.~Li, and Z.~Zhong, ``A self-adaptive feature extraction method for aerial-view geo-localization,'' \emph{IEEE Transactions on Image Processing}, 2024.

\bibitem{zhuang2021faster}
J.~Zhuang, M.~Dai, X.~Chen, and E.~Zheng, ``A faster and more effective cross-view matching method of uav and satellite images for uav geolocalization,'' \emph{Remote Sensing}, vol.~13, no.~19, p. 3979, 2021.

\bibitem{shao2023style}
J.~Shao and L.~Jiang, ``Style alignment-based dynamic observation method for uav-view geo-localization,'' \emph{IEEE Transactions on Geoscience and Remote Sensing}, vol.~61, pp. 1--14, 2023.

\bibitem{zhu2023sues}
R.~Zhu, L.~Yin, M.~Yang, F.~Wu, Y.~Yang, and W.~Hu, ``Sues-200: A multi-height multi-scene cross-view image benchmark across drone and satellite,'' \emph{IEEE Transactions on Circuits and Systems for Video Technology}, 2023.

\bibitem{workman2015wide}
S.~Workman, R.~Souvenir, and N.~Jacobs, ``Wide-area image geolocalization with aerial reference imagery,'' in \emph{Proceedings of the IEEE International Conference on Computer Vision}, 2015, pp. 3961--3969.

\bibitem{zhai2017predicting}
M.~Zhai, Z.~Bessinger, S.~Workman, and N.~Jacobs, ``Predicting ground-level scene layout from aerial imagery,'' in \emph{Proceedings of the IEEE Conference on Computer Vision and Pattern Recognition}, 2017, pp. 867--875.

\bibitem{liu2019lending}
L.~Liu and H.~Li, ``Lending orientation to neural networks for cross-view geo-localization,'' in \emph{Proceedings of the IEEE/CVF conference on computer vision and pattern recognition}, 2019, pp. 5624--5633.

\bibitem{deuser2023sample4geo}
F.~Deuser, K.~Habel, and N.~Oswald, ``Sample4geo: Hard negative sampling for cross-view geo-localisation,'' in \emph{Proceedings of the IEEE/CVF International Conference on Computer Vision}, 2023, pp. 16\,847--16\,856.

\bibitem{zhang2023cross}
X.~Zhang, X.~Li, W.~Sultani, Y.~Zhou, and S.~Wshah, ``Cross-view geo-localization via learning disentangled geometric layout correspondence,'' in \emph{Proceedings of the AAAI Conference on Artificial Intelligence}, vol.~37, no.~3, 2023, pp. 3480--3488.

\bibitem{zhu2021vigor}
S.~Zhu, T.~Yang, and C.~Chen, ``Vigor: Cross-view image geo-localization beyond one-to-one retrieval,'' in \emph{Proceedings of the IEEE/CVF Conference on Computer Vision and Pattern Recognition}, 2021, pp. 3640--3649.

\bibitem{clark2023we}
B.~Clark, A.~Kerrigan, P.~P. Kulkarni, V.~V. Cepeda, and M.~Shah, ``Where we are and what we're looking at: Query based worldwide image geo-localization using hierarchies and scenes,'' in \emph{Proceedings of the IEEE/CVF Conference on Computer Vision and Pattern Recognition}, 2023, pp. 23\,182--23\,190.

\bibitem{li2024unleashing}
G.~Li, M.~Qian, and G.-S. Xia, ``Unleashing unlabeled data: A paradigm for cross-view geo-localization,'' in \emph{Proceedings of the IEEE/CVF Conference on Computer Vision and Pattern Recognition}, 2024, pp. 16\,719--16\,729.

\bibitem{toker2021coming}
A.~Toker, Q.~Zhou, M.~Maximov, and L.~Leal-Taix{\'e}, ``Coming down to earth: Satellite-to-street view synthesis for geo-localization,'' in \emph{Proceedings of the IEEE/CVF Conference on Computer Vision and Pattern Recognition}, 2021, pp. 6488--6497.

\bibitem{shi2022geometry}
Y.~Shi, D.~Campbell, X.~Yu, and H.~Li, ``Geometry-guided street-view panorama synthesis from satellite imagery,'' \emph{IEEE Transactions on Pattern Analysis and Machine Intelligence}, vol.~44, no.~12, pp. 10\,009--10\,022, 2022.

\bibitem{li2024crossviewdiff}
W.~Li, J.~He, J.~Ye, H.~Zhong, Z.~Zheng, Z.~Huang, D.~Lin, and C.~He, ``Crossviewdiff: A cross-view diffusion model for satellite-to-street view synthesis,'' \emph{arXiv preprint arXiv:2408.14765}, 2024.

\bibitem{dai2023vision}
M.~Dai, E.~Zheng, Z.~Feng, L.~Qi, J.~Zhuang, and W.~Yang, ``Vision-based uav self-positioning in low-altitude urban environments,'' \emph{IEEE Transactions on Image Processing}, 2023.

\bibitem{ji2024game4loc}
Y.~Ji, B.~He, Z.~Tan, and L.~Wu, ``Game4loc: A uav geo-localization benchmark from game data,'' \emph{arXiv preprint arXiv:2409.16925}, 2024.

\bibitem{ye2024coarse}
Q.~Ye, J.~Luo, and Y.~Lin, ``A coarse-to-fine visual geo-localization method for gnss-denied uav with oblique-view imagery,'' \emph{ISPRS Journal of Photogrammetry and Remote Sensing}, vol. 212, pp. 306--322, 2024.

\bibitem{li2024learning}
H.~Li, C.~Xu, W.~Yang, H.~Yu, and G.-S. Xia, ``Learning cross-view visual geo-localization without ground truth,'' \emph{arXiv preprint arXiv:2403.12702}, 2024.

\bibitem{he2024contrastive}
Q.~He, A.~Xu, Y.~Zhang, Z.~Ye, W.~Zhou, R.~Xi, and Q.~Lin, ``A contrastive learning based multiview scene matching method for uav view geo-localization,'' \emph{Remote Sensing}, vol.~16, no.~16, p. 3039, 2024.

\bibitem{sun2023f3}
B.~Sun, G.~Liu, and Y.~Yuan, ``F3-net: Multi-view scene matching for drone-based geo-localization,'' \emph{IEEE Transactions on Geoscience and Remote Sensing}, 2023.

\bibitem{wang2024sequence}
Z.~Wang, D.~Shi, C.~Qiu, S.~Jin, T.~Li, Y.~Shi, Z.~Liu, and Z.~Qiao, ``Sequence matching for image-based uav-to-satellite geolocalization,'' \emph{IEEE Transactions on Geoscience and Remote Sensing}, 2024.

\bibitem{liu2024segcn}
X.~Liu, Z.~Wang, Y.~Wu, and Q.~Miao, ``Segcn: A semantic-aware graph convolutional network for uav geo-localization,'' \emph{IEEE Journal of Selected Topics in Applied Earth Observations and Remote Sensing}, 2024.

\bibitem{10587023}
Q.~Chen, T.~Wang, Z.~Yang, H.~Li, R.~Lu, Y.~Sun, B.~Zheng, and C.~Yan, ``Sdpl: Shifting-dense partition learning for uav-view geo-localization,'' \emph{IEEE Transactions on Circuits and Systems for Video Technology}, pp. 1--1, 2024.

\bibitem{tian2021uav}
X.~Tian, J.~Shao, D.~Ouyang, and H.~T. Shen, ``Uav-satellite view synthesis for cross-view geo-localization,'' \emph{IEEE Transactions on Circuits and Systems for Video Technology}, vol.~32, no.~7, pp. 4804--4815, 2021.

\bibitem{sun2024tirsa}
J.~Sun, H.~Sun, L.~Lei, K.~Ji, and G.~Kuang, ``Tirsa: A three stage approach for uav-satellite cross-view geo-localization based on self-supervised feature enhancement,'' \emph{IEEE Transactions on Circuits and Systems for Video Technology}, 2024.

\bibitem{shen2023pedestrian}
F.~Shen, X.~Shu, X.~Du, and J.~Tang, ``Pedestrian-specific bipartite-aware similarity learning for text-based person retrieval,'' in \emph{Proceedings of the 31st ACM International Conference on Multimedia}, 2023, pp. 8922--8931.

\bibitem{shen2023triplet}
F.~Shen, X.~Du, L.~Zhang, X.~Shu, and J.~Tang, ``Triplet contrastive representation learning for unsupervised vehicle re-identification,'' \emph{arXiv preprint arXiv:2301.09498}, 2023.

\bibitem{chen2019destruction}
Y.~Chen, Y.~Bai, W.~Zhang, and T.~Mei, ``Destruction and construction learning for fine-grained image recognition,'' in \emph{Proceedings of the IEEE/CVF conference on computer vision and pattern recognition}, 2019, pp. 5157--5166.

\bibitem{sun2024ultrahighresolutionsegmentationboundaryenhanced}
H.~Sun, Y.~Zhang, L.~Xu, S.~Jin, and Y.~Chen, ``Ultra-high resolution segmentation via boundary-enhanced patch-merging transformer,'' 2024.

\bibitem{cheng2024sptsequenceprompttransformer}
S.~Cheng, H.~Sun, T.~Xie, H.~Zhao, Y.~Chen, B.~Xu, and X.~Li, ``Spt: Sequence prompt transformer for interactive image segmentation,'' IEEE, pp. 1--5, 2025.

\bibitem{yin2024fine}
X.~Yin, W.~Im, D.~Min, Y.~Huo, F.~Pan, and S.-E. Yoon, ``Fine-grained background representation for weakly supervised semantic segmentation,'' \emph{IEEE Transactions on Circuits and Systems for Video Technology}, 2024.

\bibitem{ojala2002multiresolution}
T.~Ojala, M.~Pietikainen, and T.~Maenpaa, ``Multiresolution gray-scale and rotation invariant texture classification with local binary patterns,'' \emph{IEEE Transactions on pattern analysis and machine intelligence}, vol.~24, no.~7, pp. 971--987, 2002.

\bibitem{lowe1999object}
D.~G. Lowe, ``Object recognition from local scale-invariant features,'' in \emph{Proceedings of the seventh IEEE international conference on computer vision}, vol.~2.\hskip 1em plus 0.5em minus 0.4em\relax Ieee, 1999, pp. 1150--1157.

\bibitem{bouchard2005hierarchical}
G.~Bouchard and B.~Triggs, ``Hierarchical part-based visual object categorization,'' in \emph{2005 IEEE Computer Society Conference on Computer Vision and Pattern Recognition (CVPR'05)}, vol.~1.\hskip 1em plus 0.5em minus 0.4em\relax IEEE, 2005, pp. 710--715.

\bibitem{xu2018attention}
J.~Xu, R.~Zhao, F.~Zhu, H.~Wang, and W.~Ouyang, ``Attention-aware compositional network for person re-identification,'' in \emph{Proceedings of the IEEE conference on computer vision and pattern recognition}, 2018, pp. 2119--2128.

\bibitem{zhong2019invariance}
Z.~Zhong, L.~Zheng, Z.~Luo, S.~Li, and Y.~Yang, ``Invariance matters: Exemplar memory for domain adaptive person re-identification,'' in \emph{Proceedings of the IEEE/CVF conference on computer vision and pattern recognition}, 2019, pp. 598--607.

\bibitem{song2019generalizable}
J.~Song, Y.~Yang, Y.-Z. Song, T.~Xiang, and T.~M. Hospedales, ``Generalizable person re-identification by domain-invariant mapping network,'' in \emph{Proceedings of the IEEE/CVF conference on Computer Vision and Pattern Recognition}, 2019, pp. 719--728.

\bibitem{sun2018beyond}
Y.~Sun, L.~Zheng, Y.~Yang, Q.~Tian, and S.~Wang, ``Beyond part models: Person retrieval with refined part pooling (and a strong convolutional baseline),'' in \emph{Proceedings of the European conference on computer vision (ECCV)}, 2018, pp. 480--496.

\bibitem{li2017learning}
D.~Li, X.~Chen, Z.~Zhang, and K.~Huang, ``Learning deep context-aware features over body and latent parts for person re-identification,'' in \emph{Proceedings of the IEEE conference on computer vision and pattern recognition}, 2017, pp. 384--393.

\bibitem{zheng2022parameter}
Z.~Zheng, X.~Wang, N.~Zheng, and Y.~Yang, ``Parameter-efficient person re-identification in the 3d space,'' \emph{IEEE Transactions on Neural Networks and Learning Systems}, 2022.

\bibitem{shen2024stepnet}
X.~Shen, Z.~Zheng, and Y.~Yang, ``Stepnet: Spatial-temporal part-aware network for isolated sign language recognition,'' \emph{ACM Transactions on Multimedia Computing, Communications and Applications}, vol.~20, no.~7, pp. 1--19, 2024.

\bibitem{dosovitskiy2020image}
A.~Dosovitskiy, L.~Beyer, A.~Kolesnikov, D.~Weissenborn, X.~Zhai, T.~Unterthiner, M.~Dehghani, M.~Minderer, G.~Heigold, S.~Gelly \emph{et~al.}, ``An image is worth 16x16 words: Transformers for image recognition at scale,'' in \emph{International Conference on Learning Representations}, 2020.

\bibitem{he2016deep}
K.~He, X.~Zhang, S.~Ren, and J.~Sun, ``Deep residual learning for image recognition,'' in \emph{Proceedings of the IEEE conference on computer vision and pattern recognition}, 2016, pp. 770--778.

\bibitem{lu2022content}
Z.~Lu, T.~Pu, T.~Chen, and L.~Lin, ``Content-aware hierarchical representation selection for cross-view geo-localization,'' in \emph{Proceedings of the Asian Conference on Computer Vision}, 2022, pp. 4211--4224.

\bibitem{ding2020practical}
L.~Ding, J.~Zhou, L.~Meng, and Z.~Long, ``A practical cross-view image matching method between uav and satellite for uav-based geo-localization,'' \emph{Remote Sensing}, vol.~13, no.~1, p.~47, 2020.

\end{thebibliography}

\vfill

\end{document}